%% file: main.tex
\documentclass[runningheads]{llncs}

\usepackage{eccv}

\usepackage{eccvabbrv}
\usepackage{multirow}
\usepackage{multicol}

\usepackage{graphicx}
\usepackage{booktabs}

\usepackage[accsupp]{axessibility}  

\usepackage{hyperref}

\usepackage{orcidlink}

\begin{document}

\title{NVS-Adapter: Plug-and-Play Novel View Synthesis from a Single Image
} 

\titlerunning{NVS-Adapter}

\author{Yoonwoo Jeong*\inst{1}\orcidlink{0009-0004-8356-9688} \and
Jinwoo Lee*\inst{2}\orcidlink{0000-0002-9171-2016} \and
Chiheon Kim\inst{3}\orcidlink{0009-0004-6391-806X} \and \\ 
Minsu Cho\textdagger\inst{1}\orcidlink{0000-0001-7030-1958} \and
Doyup Lee\textdagger\inst{3}\orcidlink{0009-0007-7093-4991}}

\authorrunning{Jeong et al.}


\institute{POSTECH\inst{1} \,\,\,\,\,\,\, Cinamon\inst{2} \,\,\,\,\,\,\, Runway\inst{3}}

\setlength{\tabcolsep}{6.5pt}

\maketitle
\def\thefootnote{*}\footnotetext{Equally contributed.}\def\thefootnote{\arabic{footnote}}
\def\thefootnote{\textdagger}\footnotetext{Corresponding authors.}\def\thefootnote{\arabic{footnote}}

\input{preamble}
\input{sec/0_abstract}
\input{sec/1_intro}

\input{sec/2_related_work}
\input{sec/3_methods}
\input{sec/4_experiments}
\input{sec/5_conclusion}

\clearpage

\section*{Acknowledgements}
This work was supported by IITP grants (RS-2021-II212068: AI Innovation Hub (50\%), RS-2022-II220113: Developing a Sustainable Collaborative Multi-modal Lifelong Learning Framework (45\%), RS-2019-II191906: AI Graduate School Program at POSTECH (5\%)) funded by the Korea government.

\clearpage

\title{Supplementary Materials \\ NVS-Adapter: Plug-and-Play Novel View Synthesis from a Single Image
} 
\author{Yoonwoo Jeong*\inst{1}\orcidlink{0009-0004-8356-9688} \and
Jinwoo Lee*\inst{2}\orcidlink{0000-0002-9171-2016} \and
Chiheon Kim\inst{3}\orcidlink{0009-0004-6391-806X} \and \\ 
Minsu Cho\textdagger\inst{1}\orcidlink{0000-0001-7030-1958} \and
Doyup Lee\textdagger\inst{3}\orcidlink{0009-0007-7093-4991}}
\institute{POSTECH\inst{1} \,\,\,\,\,\,\, Cinamon\inst{2} \,\,\,\,\,\,\, Runway\inst{3}}
\maketitle

\input{supp_sec/1_impl_detail}
\input{supp_sec/2_controlnet}
\input{supp_sec/3_3d_consistency}
\input{supp_sec/4_ablation}
\input{supp_sec/5_num_target}
\input{supp_sec/6_concurrent}
\input{supp_sec/7_qualitative}
\input{supp_sec/8_failure_cases}
\input{supp_sec/9_figures}

\clearpage
%
%
\bibliographystyle{splncs04}
\bibliography{main}
\end{document}

%% file: preamble.tex
\newcommand{\todo}[1]{{\color{red} [#1]}}
\newcommand{\dy}[1]{{\color{blue} #1}}
\newcommand{\jw}[1]{{\color{Green} #1}}
\newcommand{\yw}[1]{{\color{teal} #1}}
\newcommand{\ms}[1]{{\color{red} #1}}
\newcommand{\ch}[1]{{\color{orange} #1}}

\newcommand{\EE}{\mathbb{E}}
\newcommand{\RR}{\mathbb{R}}

\newcommand{\cC}{\mathcal{C}}
\newcommand{\cL}{\mathcal{L}}
\newcommand{\cO}{\mathcal{O}}
\newcommand{\cV}{\mathcal{V}}
\newcommand{\cX}{\mathcal{X}}
\newcommand{\cY}{\mathcal{Y}}
\newcommand{\cZ}{\mathcal{Z}}

\newcommand{\leqsmall}{{\scriptscriptstyle\leq}}

\newcommand{\vecb}{\mathbf{b}}
\newcommand{\vecd}{\mathbf{d}}
\newcommand{\vece}{\mathbf{e}}
\newcommand{\vecf}{\mathbf{f}}
\newcommand{\vecg}{\mathbf{g}}
\newcommand{\vech}{\mathbf{h}}
\newcommand{\vecl}{\mathbf{l}}
\newcommand{\veco}{\mathbf{o}}
\newcommand{\vecp}{\mathbf{p}}
\newcommand{\vecq}{\mathbf{q}}
\newcommand{\vecr}{\mathbf{r}}
\newcommand{\vecs}{\mathbf{s}}
\newcommand{\vecu}{\mathbf{u}}
\newcommand{\vecv}{\mathbf{v}}
\newcommand{\vecx}{\mathbf{x}}
\newcommand{\vecy}{\mathbf{y}}
\newcommand{\vecz}{\mathbf{z}}
\newcommand{\vect}[1]{\boldsymbol{\mathbf{#1}}}

\newcommand{\bA}{\mathbf{A}}
\newcommand{\bM}{\mathbf{M}}
\newcommand{\bS}{\mathbf{S}}
\newcommand{\bX}{\mathbf{X}}
\newcommand{\bZ}{\mathbf{Z}}
\newcommand{\bU}{\mathbf{U}}
\newcommand{\bV}{\mathbf{V}}
\newcommand{\bW}{\mathbf{W}}

\newcommand{\sg}{\mathrm{sg}}
\newcommand{\given}{{\mkern1.5mu | \mkern1.5mu}}

\newcommand{\q}{\mathsf{q}}
\newcommand{\Q}{\mathcal{Q}}
\newcommand{\RQ}{\mathcal{R\mkern-1.5mu Q}}
\newcommand{\ARmodel}{RQ-Transformer}

\newcommand{\Nsp}{N_{\text{spatial}}}
\newcommand{\Ndep}{N_{\text{depth}}}
\newcommand{\Nnaive}{N_{\text{na\"ive}}}

\newcommand{\Lrec}{\cL_{\text{recon}}}
\newcommand{\Lcommit}{\cL_{\text{commit}}}

%% file: sec/0_abstract.tex
\begin{abstract}
Recent advancements in Novel View Synthesis (NVS) from a single image have produced impressive results by leveraging the generation capabilities of pre-trained Text-to-Image (T2I) models. However, previous NVS approaches require extra optimization to use other plug-and-play image generation modules such as ControlNet and LoRA, as they fine-tune the T2I parameters. In this study, we propose an efficient plug-and-play adaptation module, NVS-Adapter, that is compatible with existing plug-and-play modules without extensive fine-tuning. We introduce target views and reference view alignment to improve the geometric consistency of multi-view predictions. Experimental results demonstrate the compatibility of our NVS-Adapter with existing plug-and-play modules. Moreover, our NVS-Adapter shows superior performance over state-of-the-art methods on NVS benchmarks although it does not fine-tune billions of parameters of the pre-trained T2I models. The code and data are publicly available at \href{https://postech-cvlab.github.io/nvsadapter/}{postech-cvlab.github.io/nvsadapter/}
\footnote{This work was done during the summer internship program at KakaoBrain.}
\end{abstract}

%% file: sec/1_intro.tex
\section{Introduction}
\label{sec:intro}

The Novel View Synthesis (NVS) task, \ie, generating novel views from a set of images, has drawn significant attention in 3D vision due to its wide applicability in AR, VR, and robotics. Among different setups for the task, NVS from a single image is particularly challenging due to limited information in generating unseen and occluded regions of objects and scenes. 
As generative models have shown remarkable progress in learning geometry and semantics of the real world, recent methods often leverage the models as a prior to reconstruct 3D geometry from a single image, e.g., by training diffusion models~\cite{ddpm,edm,song2019generative} on NVS datasets to generate reliable multi-views~\cite{3dim,sparsefusion} or neural fields~\cite{genvs,renderdiffusion,nerfdiff}. 

Despite their notable efficacy, these methods require categorical priors of 3D objects due to the limited scale of NVS datasets, resulting in a lack of generalizability to unseen visual objects. As a breakthrough, recent work~\cite{genvs,renderdiffusion,nerfdiff} aims to improve generalizability by fine-tuning pre-trained Text-to-Image (T2I) models on large-scale 3D object datasets~\cite{objaverse,objaverse_xl}. However, since they fine-tune pre-trained T2I models, they require additional fine-tuning~\cite{zero123++} to adopt plug-and-play image generation modules such as ControlNet~\cite{controlnet} and LoRA~\cite{lora}. This fine-tuning process necessitates expensive datasets that provide both the new condition and multi-view images, limiting the usage in practice. 

In this work, we propose an efficient framework for training a plug-and-play T2I adaptation module, \emph{NVS-Adapter}, for NVS from a single image by freezing the parameters of the pre-trained T2I model. NVS-Adapter aligns the geometries of multi-views, resulting in consistent multi-view predictions that are crucial for reliable 3D generation. Moreover, our framework, which trains only additional cross-attentions, obviates the necessity of training billions of parameters, in contrast to previous approaches. In detail, our NVS-Adapter consists of two trainable components: \emph{view-consistency cross-attention} and \emph{global semantic conditioning}. The view-consistency cross-attention learns visual correspondences between views to align their local details while efficiently aggregating features from multi-views. Additionally, the global semantic conditioning facilitates the model to grasp global structure of visual objects.

Experimental results demonstrate that NVS-Adapter synthesizes geometrically consistent multi-views, achieving superior performance to previous approaches although we do not train billions of parameters of the T2I model. We further verify that NVS-Adapter is fully compatible with existing plug-and-play modules such as ControlNet~\cite{controlnet} and LoRA~\cite{lora}, allowing controlled and customized multi-view generation without additional training. 

Our contributions can be summarized as follows: 
1) We propose a novel plug-and-play NVS-Adapter with two view-alignment cross-attentions. We additionally adopt global semantic conditioning to facilitate the model to grasp the global structure of visual objects.
2) We demonstrate that NVS-Adapter generates geometrically consistent multi-views from a single image. 
3) We verify that our plug-and-play adaptation module is fully compatible with other plug-and-play T2I modules without additional training.

%% file: sec/2_related_work.tex
\section{Related Work} \label{sec:rw}

\subsubsection{Plug-and-Play Image Generation Modules} 
Existing studies propose plug-and-play image generation modules for conditional~\cite{controlnet, t2i_adapter} and customized~\cite{lora} image generation by preserving the parameters of pre-trained T2I models. T2I-Adapter~\cite{t2i_adapter} optimizes additional trainable projectors to handle auxiliary conditions. ControlNet~\cite{controlnet} clones the T2I model and optimizes the cloned one as a condition handler. LoRA~\cite{lora} efficiently optimizes residual parameters by rank decomposition matrices for customized image generation. Recently, Zero123++~\cite{zero123++} pioneers a controllable multi-view generation by fine-tuning a multi-view diffusion model with depth-conditioned ControlNet. However, it still requires an additional training process to leverage the depth-conditioned ControlNet.

\subsubsection{Generative Models for 3D Objects}
Generative models have shown remarkable progress in synthesizing 3D objects. 
Geometry-aware generative models~\cite{eg3d,giraffe,giraffehd,get3d} adopt adversarial training to implicitly learn 3D representation of faces~\cite{eg3d} or scenes~\cite{giraffe,giraffehd} from 2D images.
Recent studies train diffusion models~\cite{ddpm,edm,song2019generative} to directly generate 3D representations~\cite{3dldm,lion,ntavelis2023autodecoding,diffrf,gaudi,holodiffusion,hyperdiffusion,sdfusion,renderdiffusion,pointe,shape}.
As another direction, several work trains diffusion models on multi-view datasets~\cite{mvimgnet,co3d,shapenet} to synthesize novel views via cross-attentions~\cite{3dim,sparsefusion} or implicit feature fields~\cite{genvs,pixelnerf,nerfdiff}.\\

\subsubsection{Text-to-3D Generation via 2D Diffusion Prior} 
Recent work on T2I diffusion models enables generating 3D representations from a text prompt without 3D data. The seminal work~\cite{dreamfusion} proposes score distillation sampling (SDS) that optimizes a 3D representation by supervising its rendered views with the scores of pre-trained T2I models.
Along with the advances in neural fields~\cite{ingp,neus,neus2,dmtet}, the existing studies~\cite{sjc,magic3d,prolificdreamer,fantasia3d} have consistently improved the performance of SDS. As another direction, low-rank adaptation~\cite{lora}, structural conditions~\cite{controlnet}, and view-dependent texts~\cite{dreamfusion} are leveraged to maintain the multi-view consistency of generated 3D representations~\cite{3dfuse,realfusion,prolificdreamer}.
However, the generated 3D contents still lack geometric consistency despite their high fidelity since the training data of T2I models do not contain precise information of underlying 3D geometries. \\

\subsubsection{Fine-tuning T2I Models for Novel View Synthesis} 
Existing studies modify and fine-tune T2I models on multi-view datasets to generate high-quality and diverse images for novel view synthesis.
Zero-1-to-3~\cite{zero123,objaverse_xl} concatenates a reference image into the input of U-Net after modifying text cross-attentions for image embeddings with relative camera extrinsics~\cite{sd_img_variation}.
Despite fine-tuning on Objaverse~\cite{objaverse,objaverse_xl}, generated views suffer from geometric inconsistency and inaccurate camera viewpoints since the model cannot explicitly learn visual correspondences between the reference and generated views~\cite{zero123++}.
ConsistNet~\cite{consistnet} and SyncDreamer~\cite{syncdreamer} are built on Zero-1-to-3 to synthesize multiple novel views with an unprojection operation. MVDream~\cite{mvdream} leverages 3D self-attentions for 3D-aware multi-view generation.
Wonder3D~\cite{wonder3d} fine-tunes the Stable Diffusion Image Variations Model~\cite{sd_img_variation} to generate multi-view normal maps alongside color images.
Zero123++~\cite{zero123++} and One-2-3-45++~\cite{one2345++}, which are concurrent to our work, adopt the reference attention technique~\cite{reference_attn} and fine-tune the original self-attentions for NVS from a single image. However, they are limited to render only fixed viewpoints. MVD-Fusion~\cite{mvdfusion} jointly infers depth to produce multi-view consistent novel views. In detail, it generates multi-view RGB-D images and leverages the depth estimates to obtain reprojection-based conditioning to maintain multi-view consistency. EscherNet~\cite{eschernet} learns implicit and generative 3D representations coupled with a specialised camera positional encoding to predict novel views. Despite their capability for high-quality generation, these models still require training billions of parameters, as they fine-tune the parameters of T2I models. In contrast, our framework enables efficient training by training only additional cross-attention modules.

%% file: sec/3_methods.tex
\section{Methods} \label{sec:method}
\begin{figure*}[!t]
    \centering
    \includegraphics[width=0.95\linewidth, ]{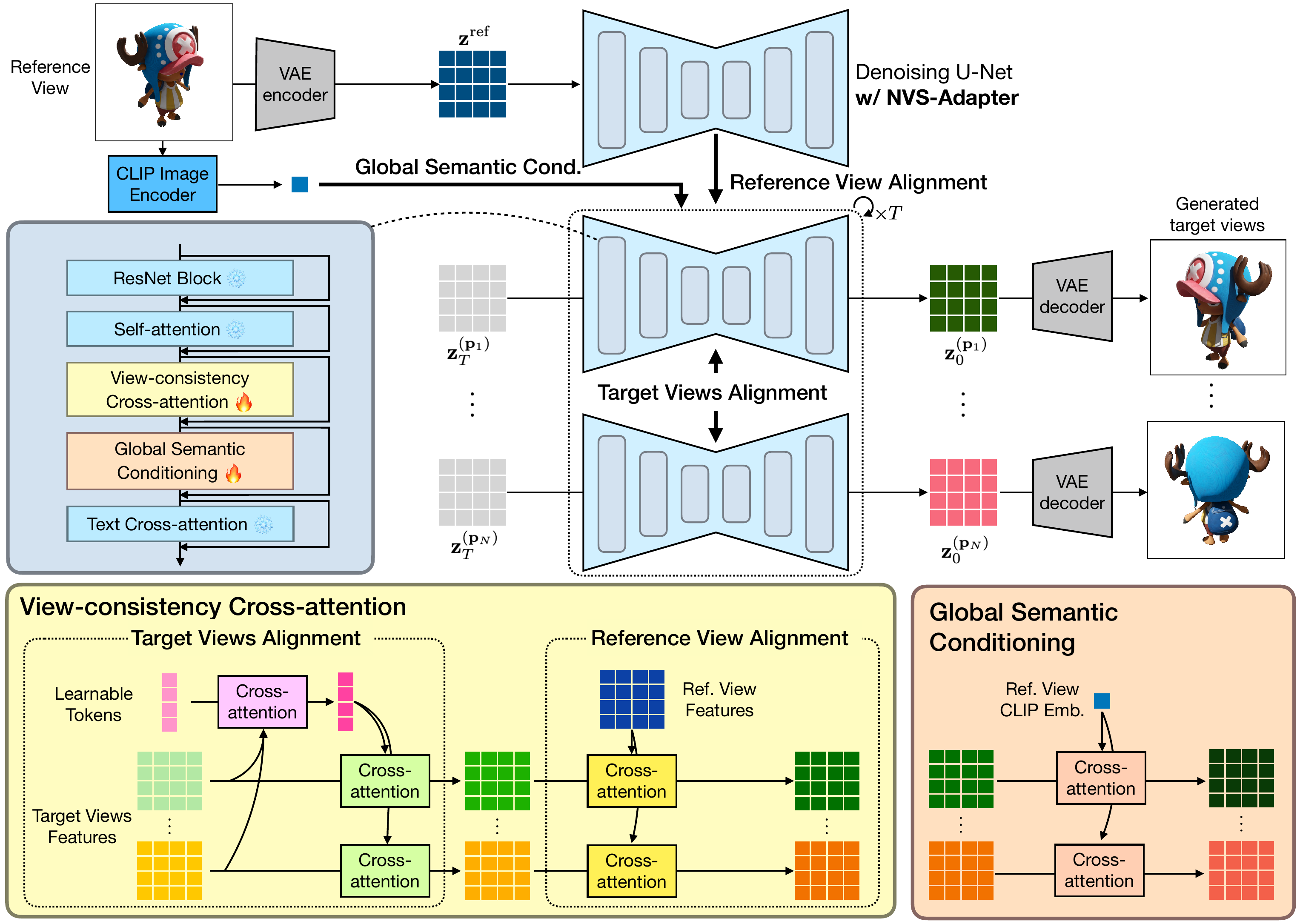}
    \caption{Overview of our framework. NVS-Adapter synthesizes novel views by incorporating two components into each U-Net block of a pre-trained T2I model: \textit{view-consistency cross-attention} which aligns features of each target view with the relevant features of other views, and \textit{global semantic conditioning} which aligns the features of target views with the global semantic structure of the reference.}
    \label{fig:framework}
\end{figure*}

Our framework synthesizes novel views from a single image for diverse 3D objects, while preserving the original parameters of T2I models. Figure~\ref{fig:framework} illustrates the overall architecture. 
We first introduce the preliminaries of diffusion models and novel view synthesis, and then formulate the task of novel view synthesis from a single image. 
Lastly, we explain how our NVS-Adapter synthesizes geometrically consistent multi-views of visual objects in a plug-and-play manner.

\subsection{Preliminaries} \label{sec:prelim}

\subsubsection{Diffusion Models}
Diffusion models learn a data distribution $q(\mathbf{x)}$ using gradually corrupted latents $\vecx_1, ..., \vecx_T$, where $q(\vecx_t | \vecx_0) := \mathcal{N}(\vecx_t; \sqrt{\alpha_t} \vecx_0, {(1-\alpha_t)} \mathbf{I})$ and $\vecx_0 = \vecx$.
Then, a diffusion model $p_\theta(\vecx_{0:T})$ aims to predict the reverse process of the Markov chain with Gaussian transitions to denoise the latents as $p_\theta(\vecx_{0:T}) := p(\vecx_T) \prod_{t=1}^{T} p_\theta(\vecx_{t-1} | \vecx_{t})$, 
where $p(\vecx_T) := \mathcal{N}(\vecx_T | \mathbf{0}, \mathbf{I})$, $p_\theta (\vecx_{t-1} | \vecx_{t}) := \mathcal{N}(\vecx_{t} ; \mu_\theta(\vecx_t, t),\sigma^2_t \mathbf{I})$, and $\sigma_t$ is a constant at $t$. 
Given a constant $\alpha_t$ at timestep $t$, the predicted mean is defined as $\mu_\theta (\vecx_{t}, t) := (\vecx_t - \sqrt{1-\alpha_t} \epsilon_\theta (\vecx_t, t)) / \sqrt{\alpha_t}$, and the noise predictor $\epsilon_\theta$ minimizes the reweighted variational lower bound~\cite{ddpm}: 
\begin{equation} \label{eq:simple_vlb}
    L_\text{simple} = \mathbb{E}_{t, \vecx_0, \epsilon}[ || \epsilon - \epsilon_\theta (\mathbf{x}_t, t) ||^2],
\end{equation}
where $\vecx_t = \sqrt{\alpha_t} \vecx_0 + \sqrt{1-\alpha_t} \epsilon$, and $\epsilon \sim \mathcal{N}(\mathbf{0}, \mathbf{I})$.

\subsubsection{Latent Diffusion Models}
Latent diffusion models (LDMs)~\cite{ldm} generate high-resolution images with their two-stage framework. In the first stage, an autoencoder~\cite{VQGAN}, which consists of an encoder $\mathcal{E}$ and a decoder $\mathcal{D}$, learns to compress a high-resolution RGB image $\vecx \in \RR^{H\times W \times 3}$ into a low-resolution feature map $\vecz = \mathcal{E} (\vecx) \in \RR^{H/f \times W/f \times C}$ with the encoder and reconstruct the image from the latent as $\vecx \approx \mathcal{D}(\mathcal{E}(\vecx))$ with the decoder, where $f$ is the downsampling factor and $C$ is the dimension of latent features.
In the second stage, the U-Net~\cite{unet,adm} of diffusion models processes features on the latent space of $\vecz$ instead of the image space of $\vecx$.
Each U-Net block of LDMs, such as Stable Diffusion, comprises a ResNet block followed by a self-attention of image features and a cross-attention between image and text features.

\subsubsection{Novel View Synthesis from a Single Image}
Given a reference view of a 3D object, we aim to generate rendered images from novel viewpoints. 
Let the reference RGB image $\vecx^\text{ref} \in \RR^{H \times W \times 3}$, $N$ relative camera viewpoints to a reference viewpoint $\mathbf{P}=\{ \mathbf{p}_1, ..., \mathbf{p}_N \}$, the rendered novel views $\vecx^\mathbf{(P)}=\{ \vecx^{(\mathbf{p}_1)}, ..., \vecx^{(\mathbf{p}_N)} \}$ are given, where $\mathbf{p}_i = [R_i; T_i] \in \RR^{3\times4}$ is $i$-th camera viewpoint with relative camera rotation $R_i \in \RR^{3\times3}$ and translation $T_i \in \RR^{3}$, and its rendered novel view $\vecx^{(\mathbf{p}_i)}$, respectively.
Then, a generative model learns a conditional distribution $q(\vecx^\mathbf{(P)} | \vecx^\text{ref}, \mathbf{P})$ to synthesize novel views $\vecx^{(\mathbf{P})}$ given a reference view $\vecx^\text{ref}$.
For the brevity of notation, we denote $q(\vecx^\mathbf{(P)} | \vecx^\text{ref}, \mathbf{P})$ by $q(\vecx^\mathbf{(P)} | \vecx^\text{ref})$.
Note that a powerful generative model such as a diffusion model is required, since the model has to extrapolate unseen or occluded parts of the 3D object. 

\subsection{LDMs for NVS from a Single Image}
We formulate the task of synthesizing novel views from a single image based on LDMs, since we exploit the pre-trained LDM, Stable Diffusion.
Specifically, a LDM learns a conditional distribution $q(\vecz^\mathbf{(P)} | \vecz^\text{ref})$,
where $\vecz^\text{ref} = \mathcal{E}(\vecx^\text{ref})$ and $\vecz^{(\mathbf{P})} = \{ \vecz^{(\mathbf{p}_i)} \}_{i=1}^{N}$ denote latent features of a reference view and $N$ target views, respectively. For simplicity, previous approaches~\cite{3dim,zero123} assume that all the target views $\vecz^{(\mathbf{P})}$ are independently drawn from the reference image, \textit{i.e.}, $p_\theta (\vecz^{(\mathbf{P})}_{t-1} | \vecz^{(\mathbf{P})}_t, \vecz^\text{ref}) = \prod_{i=1}^{N} p_\theta (\vecz^{(\mathbf{p}_i)}_{t-1} | \vecz^{(\mathbf{p}_i)}_t, \vecz^\text{ref})$, 
where noised latent features of the target views are $\vecz^{(\mathbf{p}_i)}_{t} = \sqrt{\alpha_t} \vecz^{(\mathbf{p}_i)}_{0} + \sqrt{1-\alpha_t} \epsilon$ with $\epsilon \sim \mathcal{N}(\mathbf{0}, \mathbf{I})$. Then, the approximate posterior at a specific timestep $t$ is formulated as $q(\vecz^{(\mathbf{P})}_{t} | \vecz^{(\mathbf{P})}_0, \vecz^\text{ref}) = \prod_{i=1}^{N}q(\vecz^{(\mathbf{p}_i)}_{t} | \vecz^{(\mathbf{p}_i)}_0, \vecz^\text{ref})$.
However, the independence assumption leads to geometric inconsistency across predicted target views, since the generation process does not align the geometries among the target views. 
Thus, our denoising process aligns views to consider the geometric correspondences among the target views:
\begin{equation} 
    p_\theta (\vecz^{(\mathbf{P})}_{t-1} | \vecz^{(\mathbf{P})}_t, \vecz^\text{ref}) = \prod_{i=1}^{N} p_\theta (\vecz^{(\mathbf{p}_i)}_{t-1} | \vecz^{(\mathbf{P})}_t, \vecz^\text{ref}), \label{eq:nvs_denoising}
\end{equation}
where the latent features of each $i$-th target view $\vecz_{t-1}^{(p_i)}$ is aligned with the reference view $\vecz^\text{ref}$ and the other target views $\{\vecz_{t}^{(p_j)}\}_{j \in \{1, \cdots, N\} \backslash \{i\}}$. Note that we do not corrupt the reference view to preserve fine-grained details of visual objects.

\subsection{NVS-Adapter} \label{subsec:nvs_adapter}
We propose a plug-and-play T2I adaptation module, \emph{NVS}, to effectively synthesize consistent novel multi-views from a single image.
We incorporate the adaptation modules into each U-Net block as illustrated in Figure~\ref{fig:framework}. 
While the features of different views are separately processed, the two components, \emph{view-consistency cross-attention} and \emph{global semantic conditioning}, effectively align the intermediate features across views.
The view-consistency cross-attention aligns the intermediate target features to have consistent local details by learning visual correspondences between different views.
The global semantic conditioning provide the semantic information of visual objects in the reference image.

\subsubsection{View-Consistency Cross-Attention}
\label{subsubsection:cross_view_attn}
Our view-consistency cross-attention sequentially conducts the \emph{target views alignment} and the \emph{reference view alignment}, which align a target view with the other target views and the reference view, respectively. Since every block in a U-Net share the same details, we describe the details in on one specific block without the loss of generality.

\paragraph{Target Views Alignment}
The ResNet and the self-attention block from the pre-trained T2I model separately extracts the features $\mathbf{f}^{(\mathbf{P})} = \{ \mathbf{f}^{(\mathbf{p}_i)} \}_{i=1}^{N}$ of target views, where $\mathbf{f}^{(\mathbf{p}_i)} \in \mathbb{R}^{h \times w \times c}$ is the feature of the $i$-th target view with a $h\times w$ resolution and $c$ channels. Inspired by the attention operation of Perceiver~\cite{perceiver}, we aggregate the features of the target views using $L$ learnable tokens $\mathbf{q} \in \mathbb{R}^{L \times c}$ as
\begin{equation} \label{eq:target_bottleneck}
    \bar{\mathbf{q}} := \text{MHA}(\text{Q}=\mathbf{q}, \text{KV}=\mathbf{f}^{(\mathbf{P})} + \vect{\gamma}^{(\mathbf{P})}),
\end{equation}
where Q and KV denote the query and key-value of a multi-head attention (MHA)~\cite{transformer} operation, and $\vect{\gamma}^{(\mathbf{P})}=\{ \vect{\gamma}^{(\mathbf{p}_i)} \}_{i=1}^{N}$ denotes the positional embeddings of the camera poses $\mathbf{P}$ of the target views. Note that we use relative camera poses to the reference view. When we encode the positional embeddings $\vect{\gamma}^{(\mathbf{p}_i)} \in \mathbb{R}^{h \times w \times c}$ under the $i$-th view into sinusoidal embeddings~\cite{nerf}, we represent ray positions with the Plücker coordinate~\cite{plucker} to effectively learn the geometric relation among rays.
We remark that the parameters of learnable tokens $\mathbf{q}$ are shared across all target views.
Subsequently, each target view leverages $\bar{\mathbf{q}}$ to aggregate the matched features from the other target views as
\begin{equation} \label{eq:target_view_alignment}
    \mathbf{f}_\text{TA}^{(\mathbf{P})} := \mathbf{f}^{(\mathbf{P})} + \text{MHA}(\text{Q}=
    \mathbf{f}^{(\mathbf{P})}+\vect{\gamma}^{(\mathbf{P})}, \text{KV}=\bar{\mathbf{q}}),
\end{equation}
where $\mathbf{f}_\text{TA}^{(\mathbf{P})}$ are target-views-aligned features of $\mathbf{f}^{(\mathbf{P})}$.

\paragraph{Reference View Alignment}
After the target views alignment in Eq.~\eqref{eq:target_view_alignment}, we perform the reference view alignment to ensure the consistency of details between the reference view, which is the only condition for NVS from a single image, and the generated target views. In detail, the model forwards an additional MHA operation as
\begin{equation} 
    \mathbf{f}_\text{RA}^{(\mathbf{P})} := \mathbf{f}_{\text{TA}}^{(\mathbf{P})} + \text{MHA}(\text{Q}=
    \mathbf{f}^{(\mathbf{P})}_\text{TA}+\vect{\gamma}^{(\mathbf{P})}, \text{KV}=\mathbf{f}^\text{ref}+\vect{\gamma}^\text{ref}),
\end{equation}
where $\mathbf{f}^\text{ref}$ and $\vect{\gamma}^\text{ref}$ are the ResNet feature and the positional embedding of the reference view. Similarly, the reference view alignment is applied to the reference features, considering it as a self-attention operation:
\begin{equation}
    \mathbf{f}_\text{RA}^\text{ref} := \mathbf{f}^\text{ref} + \text{MHA}(\text{Q}=
    \mathbf{f}^\text{ref}_\text{TA}+\vect{\gamma}{v}^\text{ref}, \text{KV}=\mathbf{f}^\text{ref}).
\end{equation}
The aligned features $\mathbf{f}_\text{RA}^{(\mathbf{P})}$ and $\mathbf{f}_\text{RA}^\text{ref}$ are used as an input to the next module, global semantic conditioning.

\subsubsection{Global Semantic Conditioning}
\label{subsubsection:global_condition}
Our NVS-Adapter also incorporates a cross-attention of global semantic conditioning~\cite{zero123} to inject the global semantic structure of visual objects in the reference image to different novel views.
Note that NVS from a single image assumes a reference image is given without its text captions, unlike the conventional T2I setup, which takes the text as input.
We extract CLIP~\cite{CLIP} image embeddings of the reference view to conduct a cross-attention operation between the features of each view and the CLIP image embeddings, as shown in Figure~\ref{fig:framework}.

%% file: sec/4_experiments.tex
\begin{figure}[!t]
    \centering
    \includegraphics[width=\linewidth]{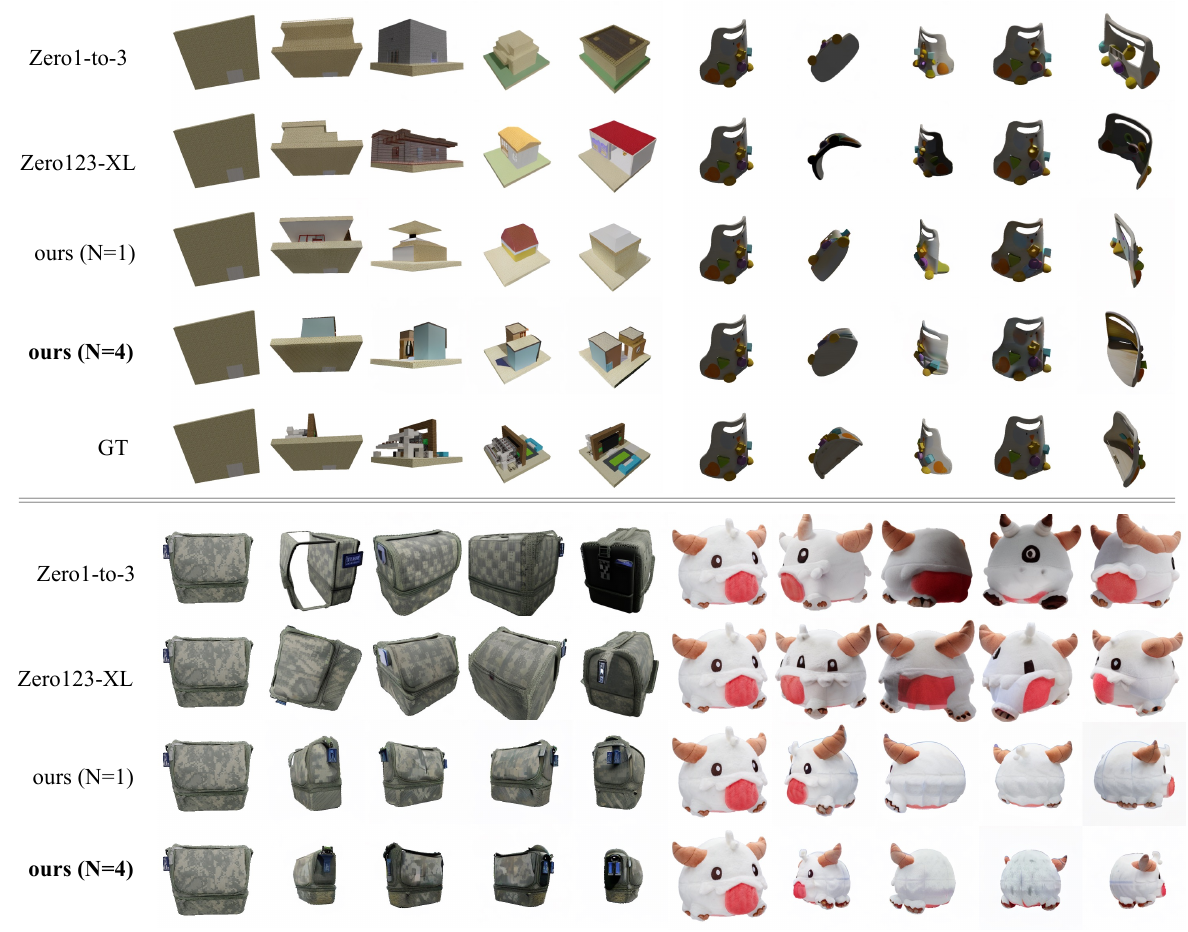}
    \caption{Novel view synthesis examples by Zero-1-to-3~\cite{zero123}, Zero123-XL~\cite{objaverse_xl}, and our NVS-Adapter with $N=1$ and $N=4$. Top images: NVS results conditioned on an image from Objaverse~\cite{objaverse} and GSO~\cite{gso} validation set. Bottom images: NVS results conditioned on a single image used in SyncDreamer~\cite{syncdreamer}. The first column presents a reference image, and the rest of four columns are synthesized views by each model. Note that the "Bottom images" do not have the ground truth images for target viewpoints since they are not from multi-view datasets.}
    \label{fig:main_nvs}
\end{figure}

\section{Experiments}
In this section, we conduct extensive experiments to demonstrate the effectiveness of our NVS-Adapter. In Section~\ref{subsection_4_1}, we evaluate our framework on NVS benchmarks and also show qualitative 3D generation results on various scenarios such as image-to-3D (I23D) and T23D. 
Then, in Section~\ref{subsection_4_2}, we show that our plug-and-play module is compatible with other plug-and-play T2I modules without extra optimization.
Furthermore, in Section~\ref{subsection_4_3}, we show our model generates high-fidelity radiance fields via Score Distillation Sampling (SDS).
Lastly, in Section~\ref{subsection_4_4} and~\ref{subsection_4_5}, we provide in-depth studies to analyze the efficacy of our NVS-Adapter to understand the geometry across different viewpoints of visual objects. 

\subsubsection{Implementation Details}
We implement our NVS-Adapter on Stable Diffusion 2.1-base~\cite{sd_v2} (SD2.1) and train our framework on Objaverse~\cite{objaverse}, which includes 800K 3D assets. To demonstrate the compatibility with ControlNet and LoRA models, we repeat this process on Stable Diffusion 1.5-base (SD1.5) instead of SD2.1, as SD1.5 provides a wider range of ControlNet~\cite{controlnet} and LoRA~\cite{lora} models and thereby allows comprehensive experimentation. For a fair comparison with Zero-1-to-3~\cite{zero123}, we use the same training dataset with the image resolution of 256$\times$256.
Our NVS-Adapter uses $h \times w$ learnable tokens $\mathbf{q}$ for target views alignment in Eq.~\eqref{eq:target_bottleneck} at every U-Net block, while aggregating $N=4$ target views.
Our model is trained for 200K iterations with the batch size of 256, using 16 NVIDIA A100 80GB GPUs for 5 days. 
For more details, please refer to the supplementary materials.

\begin{table}[!t]
\small
\caption{Comparison of NVS scores on Objaverse~\cite{objaverse} (left) and Google Scanned Objects~\cite{gso} (right) with Zero-1-to-3~\cite{zero123} and Zero123-XL~\cite{objaverse_xl}. We also report scores when our model is combined with ControlNet~\cite{controlnet} with three conditions: depth, canny, and HED. Note that no additional fine-tuning is required to attach the ControlNet. }
\centering
\resizebox{0.95\linewidth}{!}{
\begin{tabular}{l | c c c  |  c c c }
\toprule
            & \multicolumn{3}{c}{Objaverse} & \multicolumn{3}{|c}{Google Scanned Objects} \\
            \cline{2-7}
            & PSNR($\uparrow$) & SSIM($\uparrow$) & LPIPS($\downarrow$) & PSNR($\uparrow$) & SSIM($\uparrow$) & LPIPS($\downarrow$) \\
\hline
Zero-1-to-3 & 19.52 & 0.8603 & 0.1251 & 18.36 \scriptsize{$\pm$ 0.02} &  0.8418 \scriptsize{$\pm$ 0.0005} &  0.1283 \scriptsize{$\pm$ 0.0006} \\
Zero123-XL  & 17.71 & 0.8258 & 0.1552 & 18.46 \scriptsize{$\pm$ 0.11} & 0.8351 \scriptsize{$\pm$ 0.0015} & 0.1267 \scriptsize{$\pm$ 0.0009} \\ 
ours (SD21) & \textbf{19.58} & \textbf{0.8658} & \textbf{0.1135} & \textbf{18.80 \scriptsize{$\pm$ 0.07}} & \textbf{0.8469 \scriptsize{$\pm$ 0.0010}} & \textbf{0.1207 \scriptsize{$\pm$ 0.0010}}\\ \hline
ours (SD15) & 17.05 & 0.8338 & 0.1480 & 17.87 \scriptsize{$\pm$ 0.14} & 0.8331 \scriptsize{$\pm$ 0.0004} & 0.1327 \scriptsize{$\pm$ 0.0007}  \\
w. canny & 20.82 & 0.8788 & 0.0977 & 20.64 \scriptsize{$\pm$ 0.03} & 0.8624 \scriptsize{$\pm$ 0.0009} & 0.1030 \scriptsize{$\pm$ 0.0004}  \\
w. depth & 20.99 & 0.8839 & 0.0900 & 20.20 \scriptsize{$\pm$ 0.03} & 0.8611 \scriptsize{$\pm$ 0.0006} & 0.1057 \scriptsize{$\pm$ 0.0002}  \\
w. HED & \textbf{22.02} & \textbf{0.8982} & \textbf{0.0811} & \textbf{22.77 \scriptsize{$\pm$ 0.03}} & \textbf{0.8820 \scriptsize{$\pm$ 0.0003}} & \textbf{0.0912 \scriptsize{$\pm$ 0.0004}}  
\end{tabular}
}
\label{table:main_nvs_val}
\end{table}

\subsection{Novel View Synthesis from a Single Image}\label{subsection_4_1}
We evaluate our framework on the validation split of Objaverse~\cite{objaverse} and a subset of Google Scanned Objects (GSO)~\cite{gso}, containing high-quality scanned 3D assets. We use 12 and 16 rendered views for each object following the rendering pipeline of Zero-1-to-3~\cite{zero123} on Objaverse and GSO, respectively. We randomly partition 16 target views to have 4 target views for each batch and report the mean and standard deviation of results with three repetitions for GSO. Due to the huge computational budgets of the validation on Objaverse, we omit the repetition. 
Table~\ref{table:main_nvs_val} (left) shows the competitive results with previous methods of Zero-1-to-3 and Zero123-XL in terms of PSNR, SSIM, and LPIPS~\cite{lpips}. 
In Table~\ref{table:main_nvs_val} (right), our NVS-Adapter also outperforms Zero-1-to-3 and Zero123-XL on GSO, achieving higher generalization performance.
Note that our NVS-Adapter shows notably better LPIPS score than previous approaches, indicating that our model fully exploits the generation capacity of the pre-trained T2I model. We also remark that Zero-1-to-3 and Zero123-XL fine-tune over one billion parameters, and Zero123-XL learns 12 times more 3D assets than our NVS-Adapter. These results indicate our plug-and-play NVS-Adapter can efficiently and effectively adapt a T2I model to synthesize novel views from a single image.

Figure~\ref{fig:main_nvs} visualizes the synthesized novel views by our NVS-Adapter and previous methods~\cite{zero123}. Additionally, we visualize NVS results on images used in SyncDreamer~\cite{syncdreamer} to show that our NVS-Adapter well generalizes on arbitrary images that are not from multi-view datasets. 
Zero-1-to-3~\cite{zero123} and Zero123-XL~\cite{zero123, objaverse_xl} commonly suffer from the inconsistency of generated views, since they cannot synthesize multi-views at once. Although NVS-Adapter with $N=1$ lacks the consistency of generated views, it shows better correspondences between the reference and target views than Zero-1-to-3, since the view-consistency cross-attention layers learn the correspondences among view features. In contrast, our NVS-Adapter with $N=4$ well aligns the features of multi-views and effectively generates consistent multi-views at once.

\subsection{Compatibility with ControlNets and LoRAs}\label{subsection_4_2}
Since NVS-Adapter is fully compatible with existing plug-and-play modules, our framework can cover various scenarios on NVS without extra optimization. We demonstrate the effectiveness of our NVS-Adapter on two widely used scenarios, controllable generation with ControlNet~\cite{controlnet} and customized generation with LoRA~\cite{lora}, tailored to NVS from a single image. 

\begin{figure}[!t]
    \centering
    \includegraphics[width=1.0\linewidth]{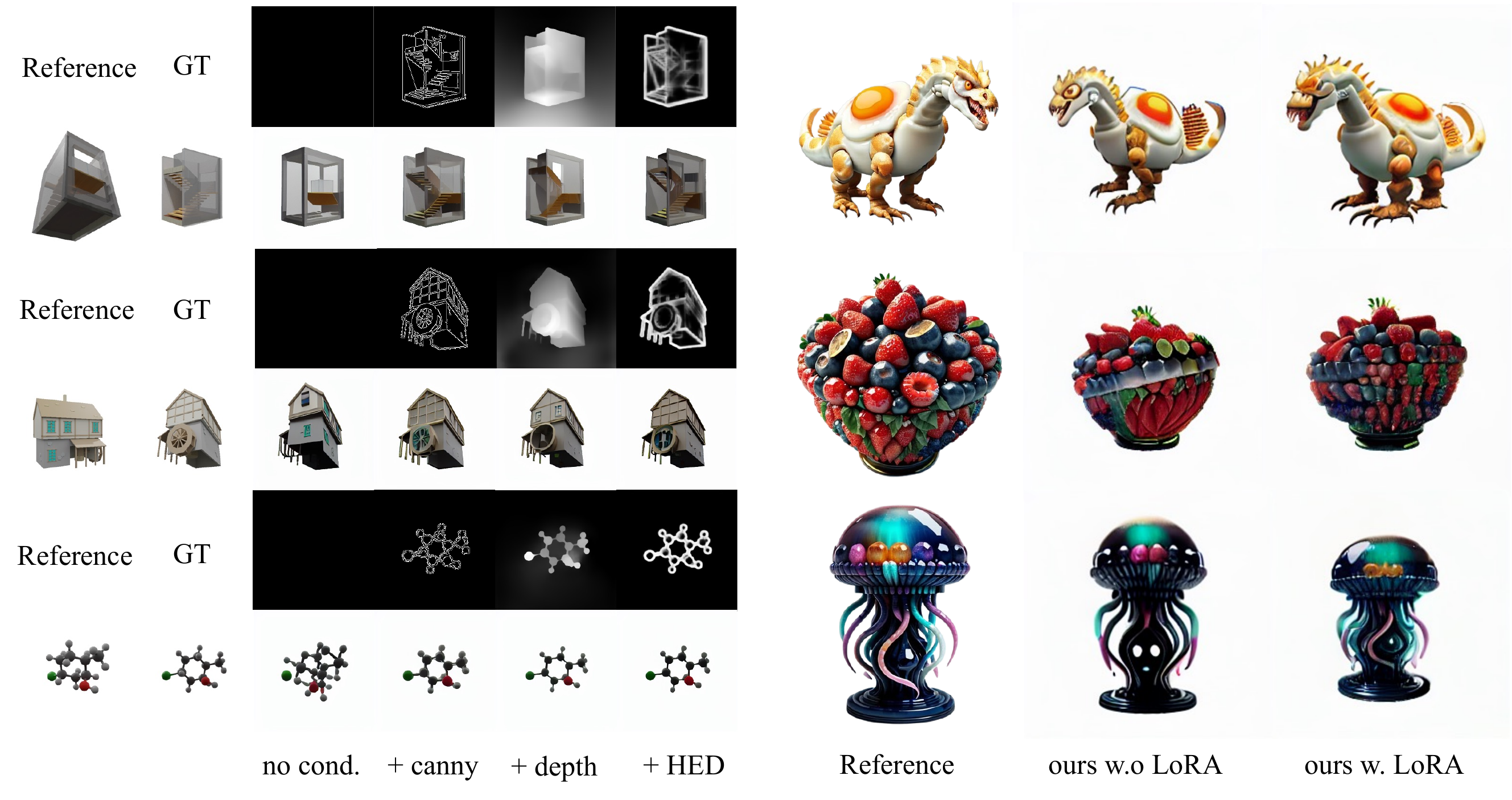}
    \caption{(left) Examples of NVS results of our NVS-Adapter with and without ControlNet~\cite{controlnet} variants. (right) Examples of NVS results of our NVS-Adapter with and without LoRA~\cite{lora} models. We use the same LoRA models for generating a reference image and multi-view images. }
     \label{fig:main_controlnet}
\end{figure}

For controllable NVS with ControlNet, we exploit the three conditions of ControlNet, canny edge, depth, and HED, to guide the multi-view generation. Table~\ref{table:main_nvs_val} shows that our NVS-Adapter can exploit canny edge, depth, and HED conditions to significantly improve the NVS performance. Figure~\ref{fig:main_controlnet} (left) shows that the synthesized novel views are guided by each condition. For instance, the first two examples are generating novel views in unobserved regions. NVS-Adapters with ControlNet address the ambiguities by the guidance from conditions. In the last example, our NVS-Adapter without ControlNet suffers from intricate geometries of a molecular toy while the ControlNet-guided NVS-Adapters generate accurate results. 

For customized NVS with LoRA, we update the pre-trained T2I model with three LoRA models -friedegg, gemstone, and blueresin- that are tailored to generate objects with unique textures. We use the same LoRA models for generating a reference image and multi-view images. We found that our NVS-Adapter with LoRA preserves the styles provided by the LoRA when generating multi-views as visualized in Figure~\ref{fig:main_controlnet} (right). According to the results, NVS-Adapter with LoRA avoids oversimplified appearance (up), captures complex details (middle), and includes physical phenomena like reflection (down). 
We remark that our framework does not require additional training for the compatibility with ControlNets and LoRAs, different from previous approaches~\cite{zero123++}. For more details, please refer to the supplementary materials.

\begin{figure}[!t]
    \centering
    \includegraphics[width=\textwidth]{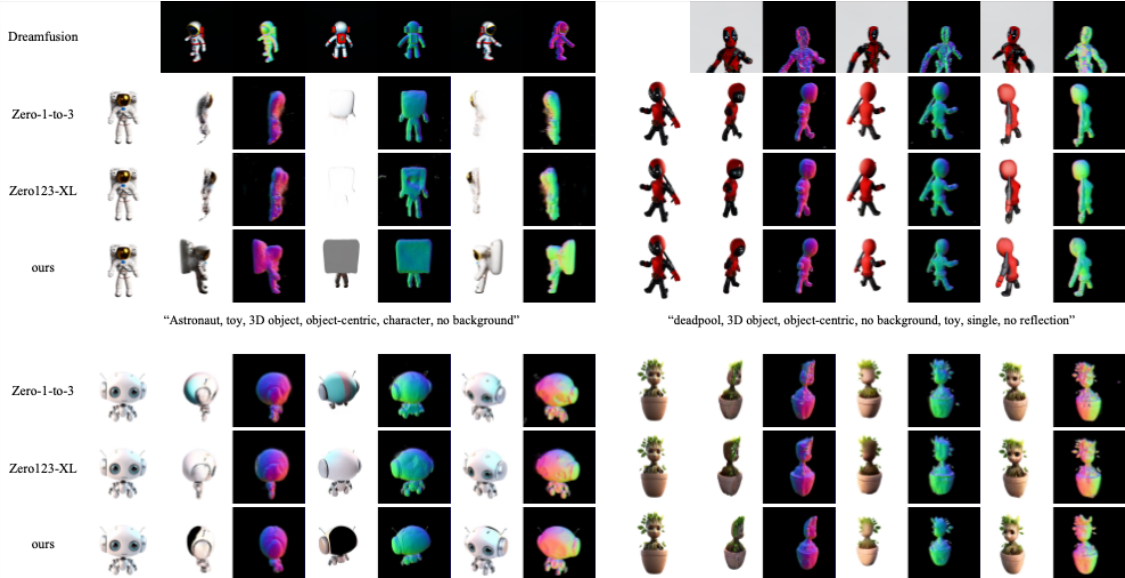}
    \caption{3D reconstruction examples via Score Distillation Sampling (SDS)~\cite{dreamfusion} with baselines~\cite{zero123,objaverse_xl} and our NVS-Adapter. Top images shows 3D reconstructions results conditioned on an image generated by T2I model, and bottom images shows results on an image in the wild.}
    \label{fig:main_qual_3d}
\end{figure}

\subsection{3D Reconstruction}\label{subsection_4_3}
We leverage our framework with NVS-Adapter to reconstruct 3D geometry from a single image, incorporating Score Distillation Sampling (SDS)~\cite{dreamfusion}.
We freeze the parameters of our framework and optimize InstantNGP~\cite{ingp} to generate 3D consistent representations.
We use threestudio~\cite{threestudio} implementations based on the setting of Zero-1-to-3 to compare our framework with other methods.
We also employ opacity regularization to reduce noisy structures.
In the coarse stage of training, we optimize the model for 5,000 iterations with batch size 16. Following the coarse stage, we further optimize the model for 3,000 iterations with a reduced batch size of 8. 
We sample a timestep $t$ in the range $[0.02, 0.98]$ and linearly decay the maximum timestep $t_{max}$ to 0.5 during the coarse stage~\cite{magic3d}. 
We attach the implementation details to the supplementary materials.

Figure~\ref{fig:main_qual_3d} compares the generated 3D representations via SDS of Zero-1-to-3, Zero123-XL, and our framework with NVS-Adapter.
We first generate images from input texts as reference views to compare our framework with Dreamfusion~\cite{dreamfusion}.
Zero-1-to-3, Zero123-XL, and our framework generate 3D following the high fidelity of images generated from texts, unlike the oversimplified results of Dreamfusion.
However, Zero-1-to-3 and Zero123-XL suffer from geometric inconsistency, but our framework shows high-quality results following the reference images.
Our framework also shows competitive results on image-to-3D, while generating geometrically consistent 3D representations compared to other methods.

\begin{table}[!t]
\small
\centering
\caption{The effects of the number of learnable tokens $L=h\times w \times c$ in target views alignment for NVS on GSO~\cite{gso}.}
\begin{tabular}{l | c c c }
\toprule
   & PSNR($\uparrow$)                      & SSIM($\uparrow$)                         & LPIPS($\downarrow$)               \\ \hline
$c$ = 0.5 & 18.26 \scriptsize{$\pm$ 0.05} & 0.8406  \scriptsize{$\pm$ 0.0010} & 0.1277  \scriptsize{$\pm$ 0.0007} \\ 
$c$ = 1 & \textbf{18.56 \scriptsize{$\pm$ 0.09}} & \textbf{0.8446  \scriptsize{$\pm$ 0.0018}} & \textbf{0.1243  \scriptsize{$\pm$ 0.0015}} \\
$c$ = 2 & 18.51 \scriptsize{$\pm$ 0.10} & 0.8441  \scriptsize{$\pm$ 0.0015} & 0.1251  \scriptsize{$\pm$ 0.0008}   \\ \bottomrule
\end{tabular}
\label{table:ablation_num_learnable}
\end{table}

\begin{table}[!t]
\centering
\small
\caption{The effects of the number of target views $N$ for NVS on GSO~\cite{gso}.}
\begin{tabular}{l | c c c}
\toprule
     & PSNR($\uparrow$)                      & SSIM($\uparrow$)                         & LPIPS($\downarrow$)                        \\ \hline
N = 1 & 14.38 \scriptsize{$\pm$ 0.02} & 0.8025 \scriptsize{$\pm$ 0.0003} & 0.1812 \scriptsize{$\pm$ 0.0013}  \\
N = 2 & 18.55 \scriptsize{$\pm$ 0.01} & 0.8451 \scriptsize{$\pm$ 0.0010} & 0.1221 \scriptsize{$\pm$ 0.0003}  \\
N = 4 & \textbf{19.01 \scriptsize{$\pm$ 0.03}} & \textbf{0.8501 \scriptsize{$\pm$ 0.0011}} & \textbf{0.1190 \scriptsize{$\pm$ 0.0013}} \\
N = 6 & 18.38 \scriptsize{$\pm$ 0.02} & 0.8419 \scriptsize{$\pm$ 0.0012} & 0.1254 \scriptsize{$\pm$ 0.0009} \\ \bottomrule
\end{tabular}
\label{table:ablation_ntarget}
\end{table}

\begin{table}[!t]
\small
\centering
\caption{The effects of using global semantic conditioning (GSC) in a view-consistency cross-attention for NVS on GSO~\cite{gso}.}
\begin{tabular}{l | c c c}
\toprule
     & PSNR($\uparrow$)                      & SSIM($\uparrow$)                         & LPIPS($\downarrow$)                        \\ \hline
ours      & \textbf{18.45  \scriptsize{$\pm$ 0.05}} & \textbf{0.8409  \scriptsize{$\pm$ 0.0003}} & \textbf{0.1240  \scriptsize{$\pm$ 0.0002}}           \\ 
\footnotesize{w/o GSC} & 18.27  \scriptsize{$\pm$ 0.05}   & 0.8388  \scriptsize{$\pm$ 0.0007} & 0.1254  \scriptsize{$\pm$ 0.0004} \\

\bottomrule
\end{tabular}
\label{table:ablation_imgemb}
\end{table}

\begin{table}[!t]
\small
\centering
\caption{The effects of changing ray positional embeddings in the view-consistency cross-attention for NVS on GSO~\cite{gso}.}
\begin{tabular}{
  l | c c c
}
\toprule
     & PSNR($\uparrow$)                      & SSIM($\uparrow$)                         & LPIPS($\downarrow$)                        \\ \hline
extrinsic & 18.02 \scriptsize{$\pm$ 0.02} & 0.8329 \scriptsize{$\pm$ 0.0006} & 0.1298 \scriptsize{$\pm$ 0.000} \\ \hline
ray o, ray d & 18.41 \scriptsize{$\pm$ 0.03} & \textbf{0.8423 \scriptsize{$\pm$ 0.0004}} & 0.1254 \scriptsize{$\pm$ 0.0004} \\ 
Plücker & \textbf{18.45 \scriptsize{$\pm$ 0.05}} & 0.8409 \scriptsize{$\pm$ 0.0003} & \textbf{0.1240 \scriptsize{$\pm$ 0.0002}} \\ 
\bottomrule
\end{tabular}
\label{table:ablation_pose}
\end{table}

\subsection{Ablation Study}\label{subsection_4_4}
We conduct an extensive ablation study to evaluate the efficacy of our NVS-Adapter in synthesizing novel views.
Due to the computational costs, we evaluate the trained models after 100K training iterations for the ablation study. For additional ablation studies, please refer to the supplementary materials. 

\subsubsection{The number of learnable tokens.}
We analyze the effects of the number of learnable tokens $\mathbf{q}$, which aggregates the features of target views in Eq.~\eqref{eq:target_bottleneck}.
With a fixed number of target views $N=4$, we vary the number of learnable tokens $L=h \cdot w \cdot c$ with $c=0.5, 1, 2$ for a $h \times w$ resolution U-Net block.
Table~\ref{table:ablation_num_learnable} shows that increasing the number of learnable tokens improves the results, since a large number of tokens can easily preserve the information of target views.
However, NVS-Adapter with $N=4$ achieves the best performance with $L=h\times w$ although it has fewer tokens than $2 \times h \times w$.
Thus, we set the number of learnable tokens to be $h\times w$ for each U-Net block for the rest of experiments. 

\subsubsection{The number of target views.}
We analyze the effects of changing the number of target views $N$. We simultaneously generate all 16 views using our models trained with $N=1,2,4,6$ to decide the number of target views to use. Since our NVS-Adapter with $N=1$ is not supervised to jointly generate multiple viewpoints, the model lacks geometric consistency when generating multiple views together. NVS-Adapter with more target views tend to show better performance, but expanding to $N=6$ deteriorates the overall performance, since the capacity of $L=h \times w$ learnable tokens is limited to aggregate $N > 4$ target views. As shown in Figure~\ref{fig:main_nvs}, our NVS-Adapter with $N=4$ consistently synthesizes multi-views in a single generation unlike our NVS-Adapter with $N=1$, which shows different geometries for each predicted viewpoint. We set $N=4$ for the rest of experiments since it achieves the best performance.

\subsubsection{The effects of global semantic conditioning.}
We conduct an ablation study to validate the effectiveness of global semantic conditioning.
Table~\ref{table:ablation_imgemb} shows that leveraging CLIP image features of a reference view enhances the semantic consistency of synthesized target views. This indicates the novel views generated without global semantic conditioning suffer from lower quality due to the lack of semantic understanding of visual objects.

\subsubsection{Ray positional embeddings.}
We examine the effects of ray positional embeddings on the view-consistency cross-attention.
We change the Plücker coordinates, which represent the ray positions in view features, into the offset and direction vectors of rays.
In addition, we also compare the results using the embedding of camera extrinsic parameters, as in Zero-1-to-3, instead of rays.
Table~\ref{table:ablation_pose} shows that utilizing each ray's position improves the performance, since it allows our view-consistency cross-attention to leverage ray geometry for matching view correspondences.

\subsection{Visualization Analysis of Cross-Attention Maps}\label{subsection_4_5}
We analyze the role of our view-consistency cross-attention to align the features of target and reference views. We visualize the mean of cross-attention maps in the last U-Net block over the attention heads. 

\begin{figure}[!t]
    \centering
    \begin{minipage}{.4\textwidth}
        \centering
        \includegraphics[width=\linewidth]{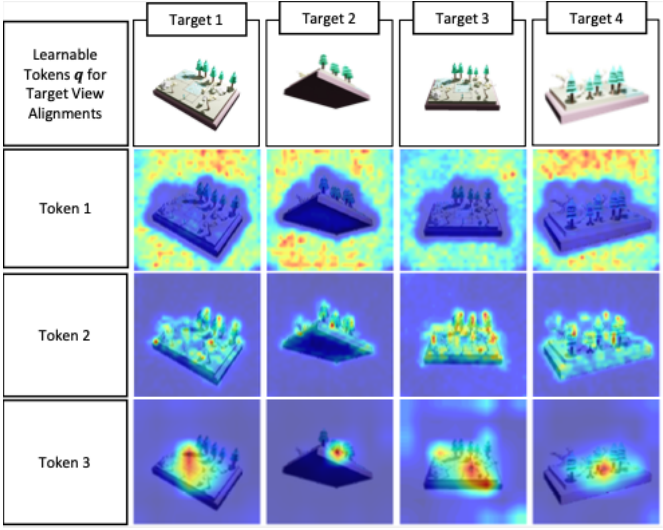}
    \end{minipage}%
    \begin{minipage}{0.6\textwidth}
        \centering
        \includegraphics[width=\linewidth]{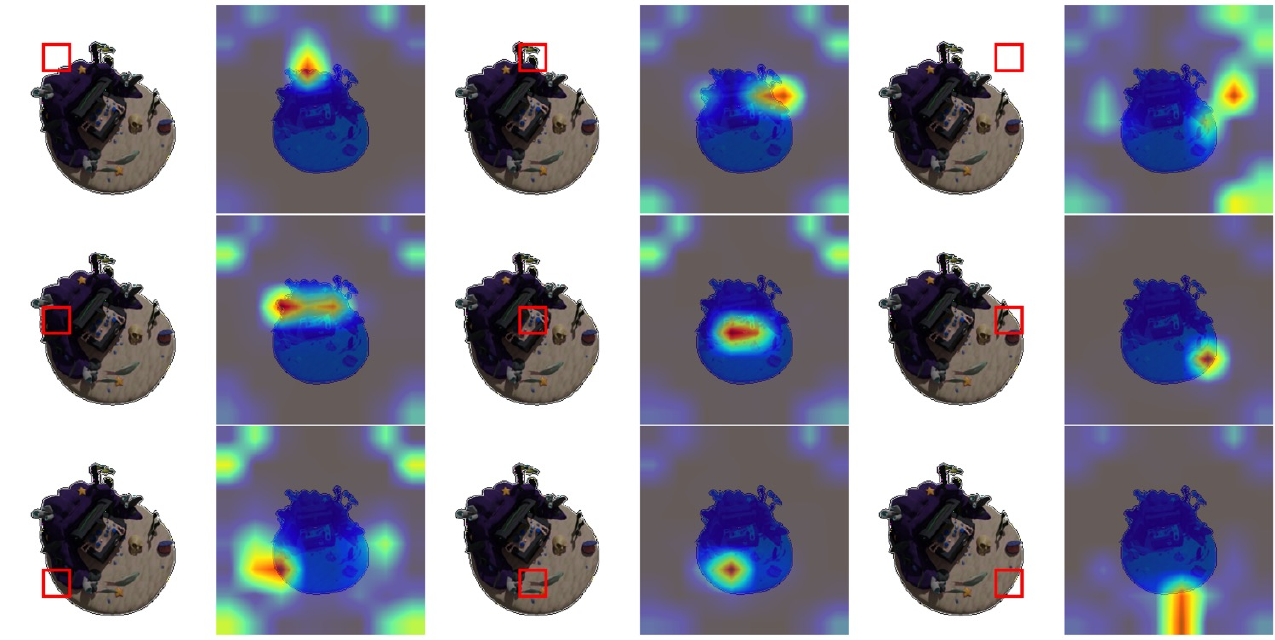}
    \end{minipage}
    \caption{Visualization of attention maps of target views alignments (left) and reference view alignments (right). For the target views alignments, each token aggregates different regions of target views. For the reference view alignments, the queries in the target view (red boxes) correspond to the highlighted regions in the reference view.}
    \label{fig:main_qual_attn_map}
\end{figure}

\subsubsection{Target views alignment.}
Figure~\ref{fig:main_qual_attn_map} (left) visualizes the cross-attention maps that use the learnable tokens as queries and the features of target views as keys.
Each learnable token effectively aggregates and encodes the information of target views based on the semantic and positional features, reducing redundancy.
As an example, token 1 (up) aggregates the white background of all target views, and token 2 (middle) attends to prominent regions in the foreground.
Meanwhile, token 3 (down) attends to a specific region of the foreground, while aggregating the same positions across different viewpoints. 
Thus, our method aggregates the features of target views into a fixed number of tokens, enabling efficient synthesis of well-aligned multi-views.

\subsubsection{Reference view alignment.}
Figure~\ref{fig:main_qual_attn_map} (right) visualizes the cross-attention maps, considering a target and reference view as a query and a key, respectively.
Note that the highlighted regions in attention maps correspond to the query regions (red box) in Figure~\ref{fig:main_qual_attn_map}.
That is, our reference view alignment focuses on matching target and reference view correspondences, aggregating the local details from the reference view for consistent synthesis.

%% file: sec/5_conclusion.tex
\section{Conclusion}
We have proposed a plug-and-play module, NVS-Adapter, for T2I models to synthesize geometrically consistent multi-views from a single image.
Our NVS-Adapter employs view-consistency cross-attention for aligning target and reference views, and the embedding of a reference view is used for the global semantic condition of views.
The experimental results show that our plug-and-play module is fully compatible with existing plug-and-play modules such as ControlNet~\cite{controlnet} and LoRA~\cite{lora}. Moreover, our NVS-Adapter achieves superior performance in NVS by learning geometric correspondences across views.

%% file: supp_sec/1_impl_detail.tex
\section{Implementation Details}

We describe the implementation details of our NVS-Adapter for reproducibility. The code and data will be publicly available. 

\subsection{Model}
We have implemented our NVS-Adapter using the framework of Stable Diffusion 2.1-base~\cite{sd_v2} (SD2.1).
We add the cross-attentions of our NVS-Adapter, where each cross-attention has the same dimensionality as the text cross-attention of each U-Net block.
We add the NVS-Adapter modules into every U-Net block of Stable Diffusion.
Our NVS-Adapter predicts $N=4$ target views at once from a single reference image. The target views alignment uses $h \times w$ learnable tokens at every U-Net block, where $h \times w$ is equal to the resolution of each U-Net feature map.
We train our NVS-Adapter on Objaverse~\cite{objaverse}, which includes 800K 3D assets, and adopt the same training data of rendered views with 256 resolution as Zero-1-to-3~\cite{zero123}.
Our model is trained for 200K iterations with 256 total batch size using 16 NVIDIA A100 80GB GPUs for 5 days, while ablation studies are conducted on the checkpoints at 100K iterations due to the limited computational resources.
A linear warmup of learning rate is used during the first 10K iterations, where the peak learning rate is 0.00005.
We use an exponential moving average (EMA) of trainable parameters with 0.9995 decay rate.
Note that the parameters of Stable Diffusion are not updated during training.
We randomly drop the information of reference images for 10\% of training samples to apply classifier-free guidance (CFG)~\cite{cfg} during view generation.
To drop the information of reference images, we replace the global semantic conditioning and positional embeddings with zero embeddings, and replace the U-Net input of reference images with Gaussian noises.
We use 11.0 scale of CFG during inference.
The codes and evaluation samples will be released to ensure the reproducibility of our results.

\subsection{ControlNet and LoRA Experiments}
We additionally train our NVS-Adapter with Stable Diffusion 1.5-base~\cite{sd_v1} (SD1.5) to show compatibility with ControlNet~\cite{controlnet} and LoRA~\cite{lora} modules. Here, we elaborate thorough details for ControlNet and LoRA experiments. 

For ControlNet experiments, we use three publicly available modules with three conditions: depth, canny edge~\cite{canny}, and HED edge~\cite{hed}. For depth conditions, we use a pretrained DPT~\cite{dpt} model to extract the monocular depths from target images and then we normalize the extracted depth values from 0 to 1. As suggested in~\cite{controlnet}, we extract canny edges with the low and high threshold of 100 and 200, respectively.

For LoRA experiments, we select seven modules that are suitable for multi-view generation and compatible with SD1.5. We list the links in footnotes. \footnote{Friedegg: \url{https://civitai.com/models/255828?modelVersionId=288399}}, \footnote{Natural gemstone: \url{https://civitai.com/models/273611?modelVersionId=308389}}, \footnote{Blueresin: \url{https://civitai.com/models/256890?modelVersionId=289671}}, \footnote{Liquid flow: \url{https://civitai.com/models/199968?modelVersionId=225007}}, \footnote{Chocolate: \url{https://civitai.com/models/197998?modelVersionId=222742}}, \footnote{Wood: \url{https://civitai.com/models/185218?modelVersionId=239739}}, and \footnote{Gelato: \url{https://civitai.com/models/67122?modelVersionId=264333}}.
To inject LoRA models in our NVS-Adapter, we load the LoRA-injected SD1.5 weights instead of the vanilla SD1.5 checkpoint. We use the same LoRA model for generating reference images and generating multi-view images.

\subsection{3D Reconstruction}
We have implemented Score Distillation Sampling (SDS)~\cite{dreamfusion} by building NVS-Adapter on threestudio~\cite{threestudio} for 3D reconstruction. We used the CFG scale of 3.0 for sampling. The distance from cameras to the center of the scene and the vertical field of view (FoV) are fixed to 1.5 and 49.1 degrees, respectively. We sample elevations from the range of [-10, 80] and azimuths from the range of [-180, 180]. The rendering resolution starts at 64 $\times$ 64 in the coarse stage and is increased to 256 $\times$ 256 in the fine stage.
We optimize the 3D model using AdamW~\cite{AdamW} with the learning rate of 0.01.

%% file: supp_sec/2_controlnet.tex
\section{ControlNet with Multiple Conditions}

We check whether the rendering quality of NVS-Adapter with ControlNet is further improved when multiple conditions are jointly used. Table~\ref{tab:reb_3} also investigates the effect of leveraging multiple conditions jointly, showing NVS scores on varying conditions. Since HED provides sufficiently strong supervision for guidance, using HED conditions only has achieved the best rendering quality.

\begin{table}[h]
    \centering
    \vspace{-3mm}
    \caption{Multi-condition experiments with ControlNet.}
    \vspace{-2mm}
    \label{tab:reb_3}
    \resizebox{0.6\linewidth}{!}{
    \begin{tabular}{ccc|ccc}
          HED & Depth & Canny & PSNR & SSIM & LPIPS \\
        \hline
         & & & 17.87 & 0.8331 & 0.1327  \\
         & & \checkmark & 20.64 & 0.8624 & 0.1030 \\
         & \checkmark & & 20.20 & 0.8611 & 0.1057 \\
         \checkmark & & & 22.77 & 0.8820 & 0.0912 \\
         & \checkmark & \checkmark & 18.90 & 0.7697 & 0.1269  \\
         \checkmark & \checkmark & & 19.64 & 0.7757 & 0.1254 \\
         \checkmark & & \checkmark & 19.80 & 0.7879 & 0.1236 \\
         \checkmark & \checkmark & \checkmark & 19.91 & 0.7842 & 0.1222 \\
    \end{tabular}
    }
    \vspace{-3mm}
\end{table}

%% file: supp_sec/3_3d_consistency.tex
\section{3D Consistency Score}

To demonstrate that our model shows superior 3D consistency to previous models, we quantitatively compare the consistency of rendered images. In detail, following the seminal work, DreamFusion~\cite{dreamfusion}, we evaluate clip similarity (CLIP-S) of rendered images with an input condition such as a text prompt for the text-to-3D task (T23D) or a reference image for the image-to-3D task (I23D). In Table~\ref{tab:reb_2}, NVS-Adapter achieves superior CLIP-S for the text-to-3D task and competitive CLIP-S for the image-to-3D task.

\begin{table}
    \centering
    \caption{3D consistency(CLIP-S) score comparison.}
    \label{tab:reb_2}
    \begin{tabular}{c|c|c}
         CLIP-S & T23D & I23D \\
        \hline
        DreamFusion & 0.2920 & -   \\
        Zero123 & 0.2910 & \textbf{0.8623} \\
        Zero123-XL & 0.2974 & \textbf{0.8618} \\
        NVS-Adapter & \textbf{0.2993} & \textbf{0.8621}
        \\
    \end{tabular}
\end{table}

%% file: supp_sec/4_ablation.tex
\section{Ablation Studies}

We conduct additional ablation studies for thorough analysis of NVS-Adapter. 

\subsubsection{Full cross-attention architecture} 
We conduct an ablation study for learnable tokens. In detail, we use full cross-attention instead of using learnable tokens. As shown in Table~\ref{tab:ablation_full}, ours with learnable tokens shows competitive performance with the full cross-attention model. Note that full cross-attention has $O(N^2)$ complexity and our attention has $O(N)$ complexity where $N$ denotes the number of queries.

\subsubsection{The effects of full fine-tuning.} 

In Table~\ref{tab:ablation_full}, we evaluate the performance of our framework after training all parameters, including the T2I model. For sufficient training, we train our NVS-Adapter and full fine-tuned models for 200K iterations. 
Full fine-tuning of the model outperforms our plug-and-play approach, since full fine-tuning updates over one billion parameters tailored to NVS.
However, our framework shows competitive results compared to the fully fine-tuned model, while preserving the original capability of T2I models.
These results emphasize the effectiveness of our plug-and-play NVS-Adapter, enabling T2I models to generate multi-views from images or texts, along with T2I generation.

\begin{table}[!t]
    \centering
    \vspace{0mm}
    \caption{NVS-Adapter ablation studies.}
    \vspace{-2mm}
    \label{tab:ablation_full}
    \begin{tabular}{c|ccccc}
         & Time (h) & Memory (GB) & PSNR & SSIM & LPIPS \\
        \hline 
         Zero123 - & - & 18.34 & 0.8424 & 0.1283 \\
        \hline
         ours & \textbf{71.0} & \textbf{28.80} & 18.87 & 0.8477 & 0.1185  \\
         full fine-tune & 104.3 & 35.11 & \textbf{19.00} & \textbf{0.8488} & \textbf{0.1151} \\
         full cross-attn &  96.2 & 32.29 & \textbf{18.93} & 0.8468 & 0.1195 \\
    \end{tabular}
    \vspace{-5mm}
\end{table}

%% file: supp_sec/5_num_target.tex
\section{Generalizing the Number of Target Views}

We evaluate the changes of the number of target views, considering that our NVS-Adapter is flexible to generalize the number of target views.
Specifically, after a framework with NVS-Adapter is trained to synthesize $N_\text{train}=\{ 1, 2, 4, 6\}$ target views, $N_\text{eval}=\{1, 2, 4, 6, 16 \}$ target views are generated at once during inference.
Table~\ref{table:supp_ntarget} shows PSNR, SSIM, and LPIPS scores on GSO~\cite{gso} according to different pairs of ($N_\text{train}$, $N_\text{eval}$). 
When our framework with NVS-Adapter is trained to synthesize a single target view as $N_\text{train}=1$, the results are competitive with other settings.
However, the framework with $N_\text{train}=1$ fails to synthesize multiple views at once and cannot generalize to increase the number of target views for consistent novel view synthesis.
However, when our frameworks with $N_\text{train}=\{ 2, 4, 6 \}$ are trained to synthesize multiple views at once, our frameworks can increase the number of target views.
The NVS-Adapter with $N_\text{train=4}$ shows well-balanced results on various number of target views, while achieving its best performance on $N_\text{eval}=16$. 
Thus, we have finally selected $N_\text{train}=4$ as our main hyper-parameter setup, considering its generalization capability.

\begin{table*}[!t]
\small
\centering
\caption{The effects of the number of target views $N$ on GSO. The first four rows report PSNR scores, the next four rows report SSIM, and the last four rows report LPIPS, with varying the number of target views for training and evaluation. }
\begin{tabular}{c | c|c c c c c c}
\toprule
&  & $N_{\text{eval}} = 1$ &  $N_{\text{eval}} = 2$   & $N_{\text{eval}} = 4$ &  $N_{\text{eval}} = 6$ &  $N_{\text{eval}} = 16$  \\
 \hline
\multirow{4}{*}{PSNR} & $N_{\text{train}}$ = 1 & \textbf{18.45 } &  15.09  &  14.58  &  14.50 &  14.38  \\
& $N_{\text{train}}$ = 2 &  18.42  & \textbf{18.70 } &  \textbf{18.62 } &  \textbf{18.67 } &  18.55  \\
& $N_{\text{train}}$ = 4 &  17.08  &  18.01  & 18.56  &  \textbf{18.67 } &  \textbf{19.01 } \\
& $N_{\text{train}}$ = 6 &  15.09  &  16.23  &  17.44  & 17.85  &  18.38  \\

\hline

\multirow{4}{*}{SSIM} & $N_{\text{train}}$ = 1 & 0.8409  &  0.8057  &  0.8013  &  0.8013 &  0.8025  \\
& $N_{\text{train}}$ = 2 &  \textbf{0.8427 } & \textbf{0.8453 } &  \textbf{0.8450 } &  0.8453  &  0.8451  \\
& $N_{\text{train}}$ = 4 &  0.8237  &  0.8374  & 0.8446  &  \textbf{0.8456 } &  \textbf{0.8501 } \\
& $N_{\text{train}}$ = 6 &  0.7924  & 0.8126  &  0.8288  & 0.8333  & 0.8419  \\

\hline

\multirow{4}{*}{LPIPS} & $N_{\text{train}}$ = 1 & 0.1240  &  0.1706  & 0.1800  &  0.1805  & 0.1812  \\
& $N_{\text{train}}$ = 2 &  \textbf{0.1239 } & \textbf{0.1215  }& \textbf{0.1219 } &  \textbf{0.1217 } &  0.1221  \\
& $N_{\text{train}}$ = 4 &  0.1460  &  0.1307  & 0.1243  &  0.1228  & \textbf{ 0.1190 } \\
& $N_{\text{train}}$ = 6 &  0.1796  &  0.1581  & 0.1397  & 0.1335  &  0.1254  \\

\bottomrule
\end{tabular}
\label{table:supp_ntarget}
\end{table*}

%% file: supp_sec/6_concurrent.tex
\section{Comparisons of NVS-Adapter with Concurrent Work}

We compare the quantitative and qualitative results of our NVS-Adapter with our concurrent projects such as SyncDreamer~\cite{syncdreamer} and Zero123++~\cite{zero123++}, which synthesize multiple novel views at once. 
We compare NVS-Adapter with recent multi-view generation methods, SyncDreamer and Zero123++. Considering their rendering is limited only to the predefined viewpoints, we evaluate our method on rendering their fixed viewpoints on GSO. 
We exclude One-2345 in comparison, since it relies on the frozen Zero123 for NVS and is also compatible with NVS-Adapter. 
Our model achieves competitive performance with SyncDreamer in Table~\ref{tab:reb_1}, although our model is not trained with fixed viewpoints.
In contrast, NVS-Adapter underperforms Zero123++ under its predefined viewpoints. Considering that Zero123++ adopts the shift of noise schedule and progressive training of full parameters with unknown training iterations, exploring the training strategy would be worth exploration, while keeping our modularity of NVS-Adapter and arbitrary viewpoint prediction.

\begin{table}
    \centering
    \caption{NVS score comparison}
    \label{tab:reb_1}
    \begin{tabular}{c|c|c|c}
          & PSNR & SSIM & LPIPS \\
        \hline
        SyncDreamer & \textbf{19.19} & 0.9050 & 0.1099 \\
        NVS-Adapter & 18.75 & \textbf{0.9197} & \textbf{0.0917} \\
        \hline
        Zero123++ & \textbf{21.64} & \textbf{0.9333} & \textbf{0.0847} \\
        NVS-Adapter & 18.27 & 0.9141 & 0.0958 \\
    \end{tabular}
\end{table}

Figure~\ref{fig:qual_syncdreamer} and Figure~\ref{fig:qual_zero123plus} represent the synthesized novel views to compare our NVS-Adapter with SyncDreamer and  Zero123++ respectively.
Although our NVS-Adapter preserves the original parameters of T2I models, our frameworks synthesize geometrically consistent multi-views, demonstrating competitive results with SyncDreamer and Zero123++. 

We have also considered MVDream~\cite{mvdream} as our baseline model. Both MVDream and NVS-Adapter jointly generate multiple novel views with geometric consistency by attention layers. Specifically, MVDream applies 3D attention having all pixels in multi-views as query, key, and value, but NVS-Adapter uses learnable tokens for better efficiency. In addition, MVDream uses camera embeddings to render fixed viewpoints whereas NVS-Adapter uses ray embeddings to render arbitrary viewpoints. However, MVDream uses a text condition to generate its multi-views whereas NVS-Adapter uses a reference view to generate the aligned and novel multi-views. Therefore, we finally have excluded MVDream from our baselines.

%% file: supp_sec/7_qualitative.tex
\section{More Qualitative Results}

We attach more qualitative results in the main paper.
Figure~\ref{fig:qual_controlnet} visualizes additional examples of NVS results of NVS-Adapter with and without ControlNet modules.
Figure~\ref{fig:qual_lora} visualizes additional examples of NVS results of NVS-Adapter with and without LoRA modules.
Figure~\ref{fig:supp_qual_objaverse} and Figure~\ref{fig:supp_qual_gso} show the examples of generated novel views on Objaverse~\cite{objaverse} and GSO~\cite{gso}, respectively.
Figure~\ref{fig:supp_qual_inthewild} shows the results of generated views from arbitrary reference images, which are collected on the internet.
Figure~\ref{fig:supp_qual_textprompt1} and Figure~\ref{fig:supp_qual_textprompt2} show the examples of text to multi-views, where the reference imges are generated by the T2I model.
Figure~\ref{fig:supp_qual_sds_text} and Figure~\ref{fig:supp_qual_sds_image} show additional results on score distillation sampling(SDS)~\cite{dreamfusion} to generate 3D objects. We also attach videos for the generated 3D.
Figure~\ref{fig:supp_qual_target_attn} visualizes attention maps of target view alignments, which aggregate the information of target views into learnable tokens.

%% file: supp_sec/8_failure_cases.tex
\section{Failure Cases}

Figure~\ref{fig:supp_qual_failure} shows the failure cases of our model. 
The first example shows the case where NVS-Adapter misunderstands the direction of cars although the rider (bart) is reliably rendered. 
The second example assumes the object to be a thin-plane object. 
Since Objaverse includes numerous thin-plane objects, the model can result in volumeless generation. 
In the case of the third example, the doll has inconsistent color for the unobserved parts in a reference image. 
Although unobserved parts are ambiguous, the synthesized colors do not make sense in general. 
The color of the back should be purple rather than yellow. 
The fourth example seems strange since the ghost is located above the dish, indicating wrong topology between the ghost and the dish.
The last example includes complex backgrounds. NVS-Adapter fails to reliably generate backgrounds since NVS-Adapter haven't been observed scenes with backgrounds.
We believe that increasing the scale of the training dataset can improve the performance of our NVS-Adapter. 

Although our framework can synthesize consistent multiple novel views of diverse objects, our framework still relies on SDS with a large diffusion model that requires per-scene optimization for generating a 3D representation.
We expect that increasing the number of target views enables a neural field to be directly optimized via the generated multi-views, reducing the generation time of neural fields.
Exploiting generalizable neural fields~\cite{ipc,locality_inr,one2345++} is worth an exploration to circumvent the need for explicit optimization of neural fields per each sample.
Furthermore, generating novel views of natural scenes from a single image is an interesting future work, expanding beyond visual objects.

%% file: supp_sec/9_figures.tex
\begin{figure*}
    \centering
    \includegraphics[width=0.7\linewidth]{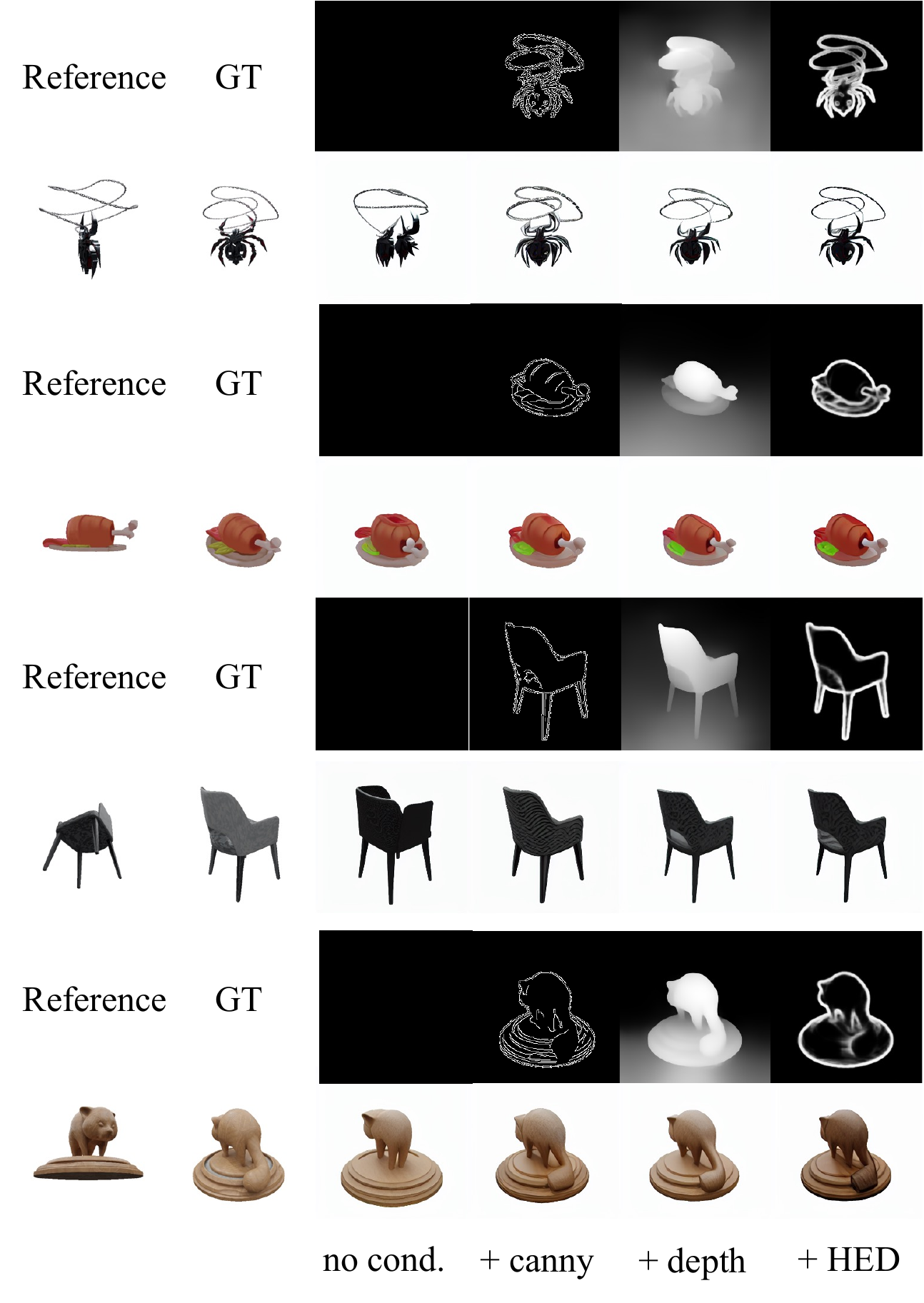}
    \caption{ Examples of NVS results of our NVS-Adapter with and without ControlNet~\cite{controlnet} variants. }
     \label{fig:qual_controlnet}
\end{figure*}

\begin{figure*}
    \centering
    \includegraphics[width=1.0\linewidth]{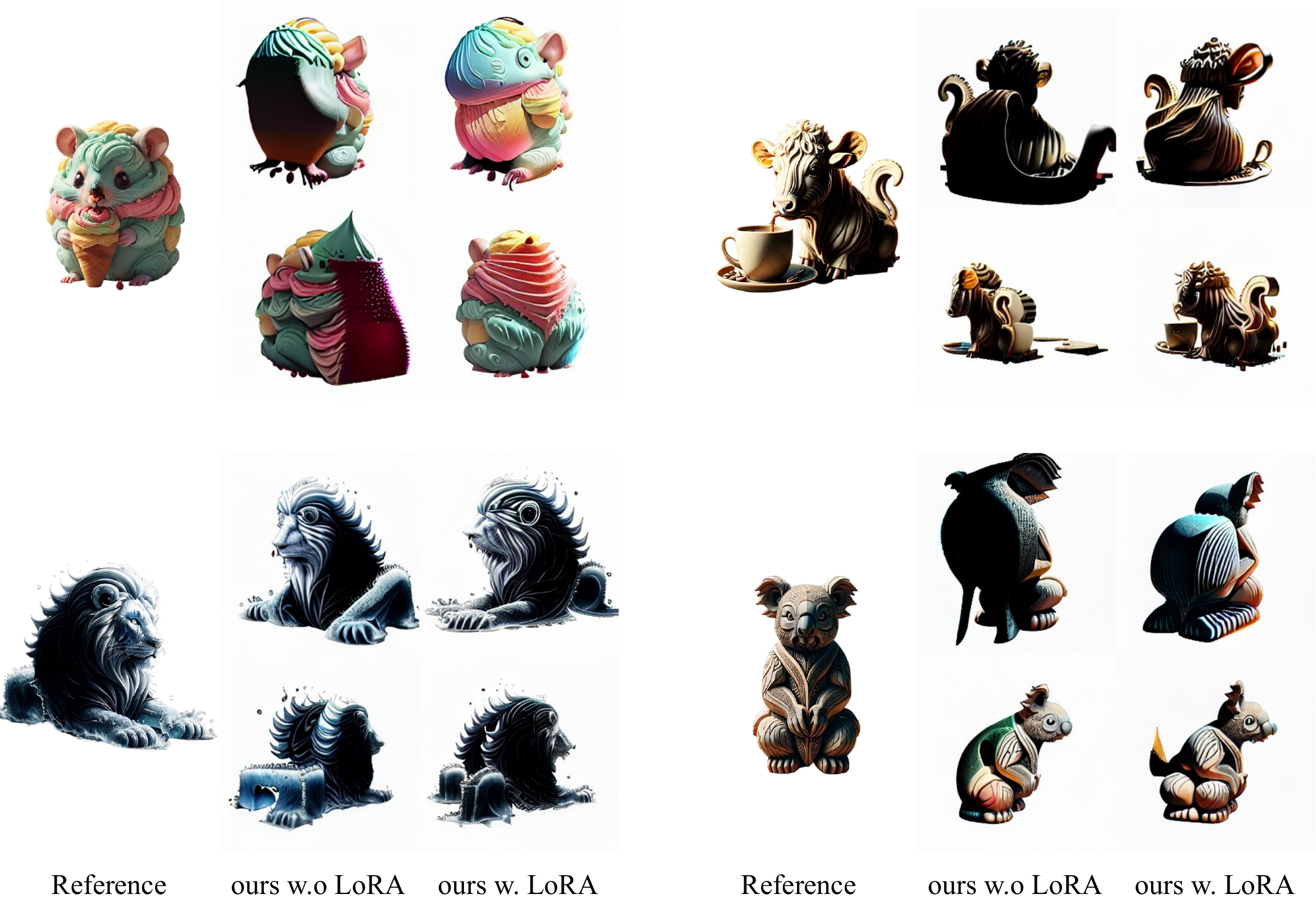}
    \caption{ Examples of NVS results of our NVS-Adapter with and without LoRA~\cite{controlnet} modules. }
     \label{fig:qual_lora}
\end{figure*}

\begin{figure*}
    \centering
    \includegraphics[width=\textwidth]{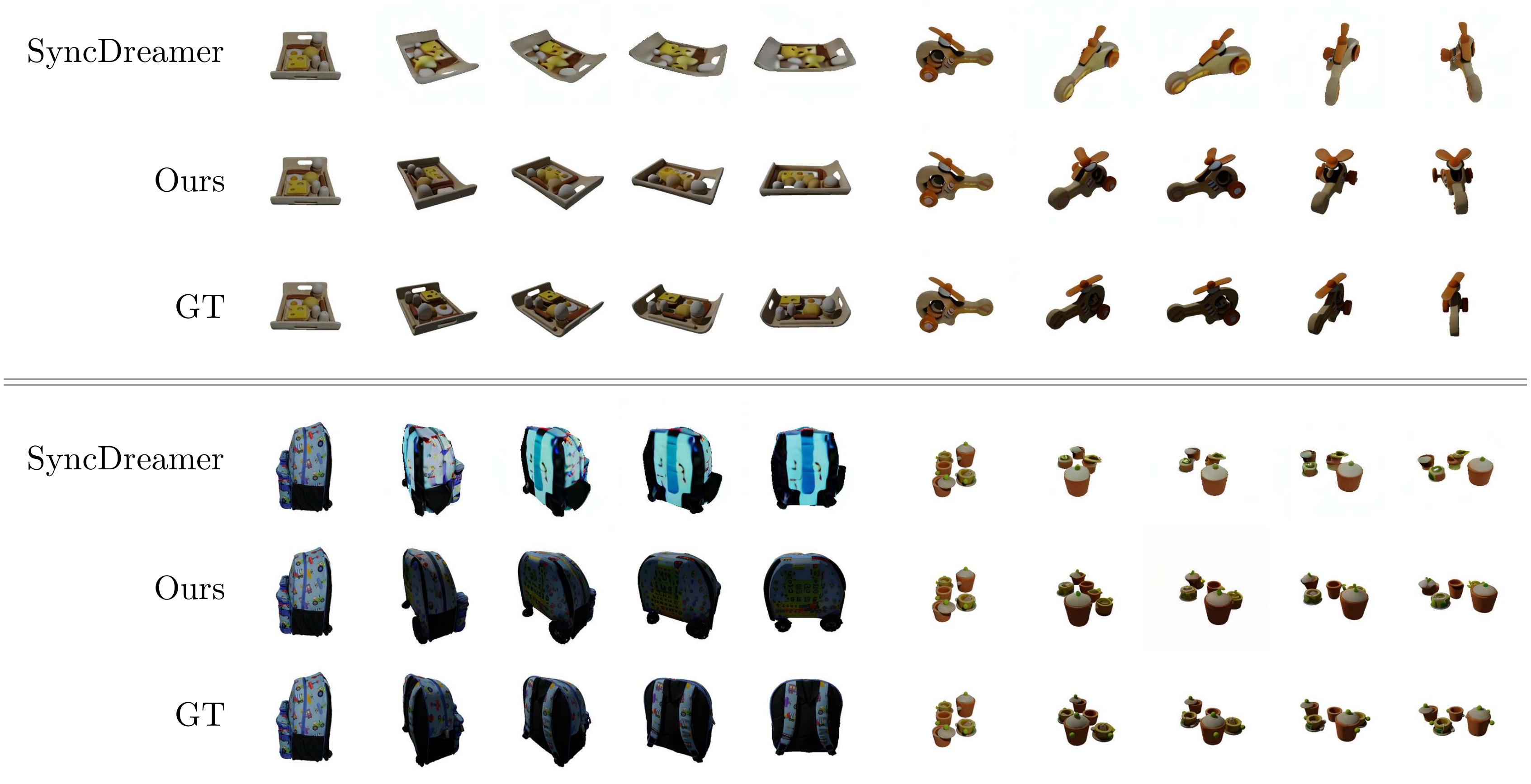}
    \caption{Novel view synthesis examples by SyncDreamer and our NVS-Adapter.}
    \label{fig:qual_syncdreamer}
    \vspace{-0.1in}
\end{figure*}

\begin{figure*}
    \centering
    \includegraphics[width=\textwidth]{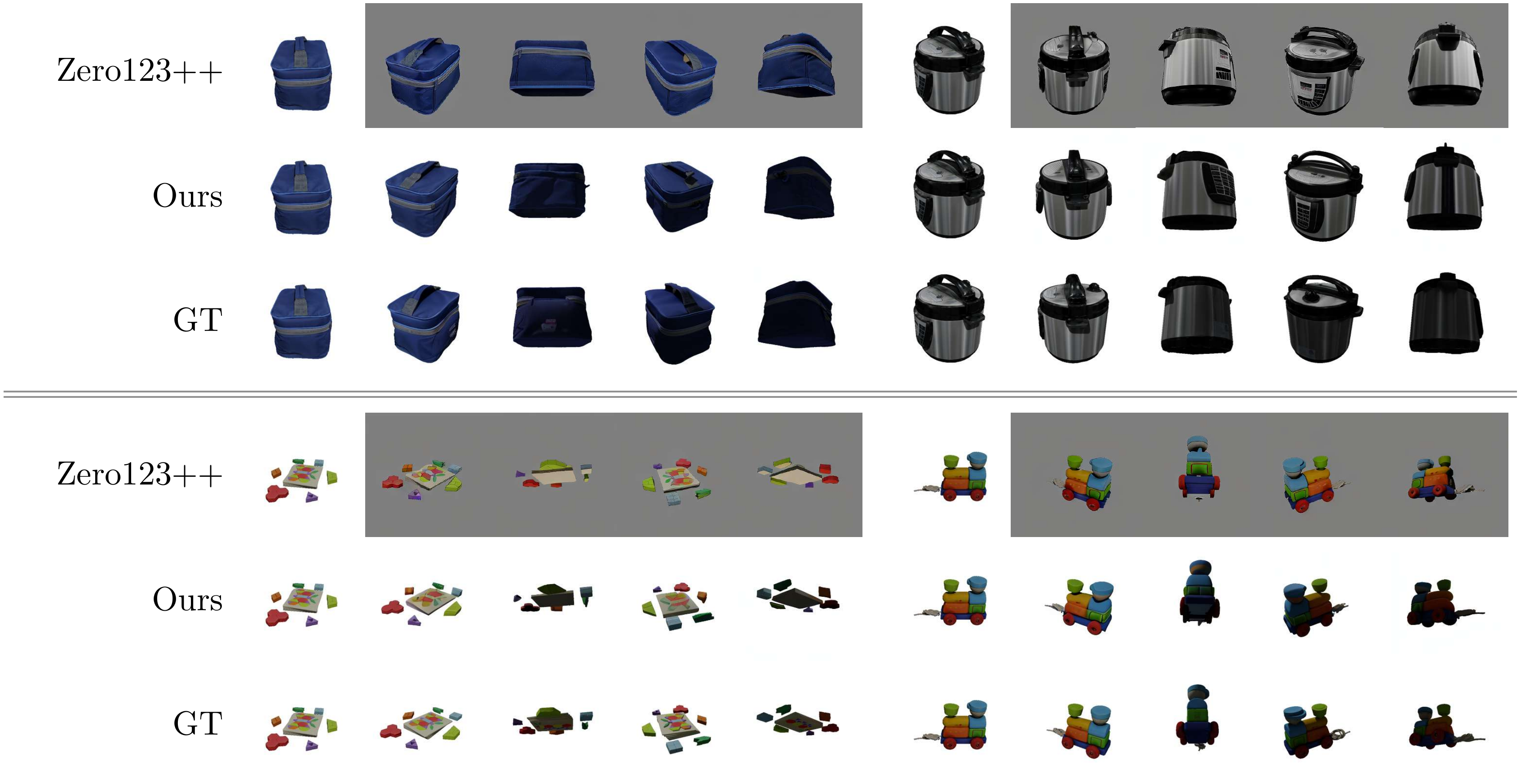}
    \caption{Novel view synthesis examples by Zero123++ and our NVS-Adapter.}
    \label{fig:qual_zero123plus}
    \vspace{-0.1in}
\end{figure*}

\begin{figure}
    \centering
    \includegraphics[width=\textwidth]{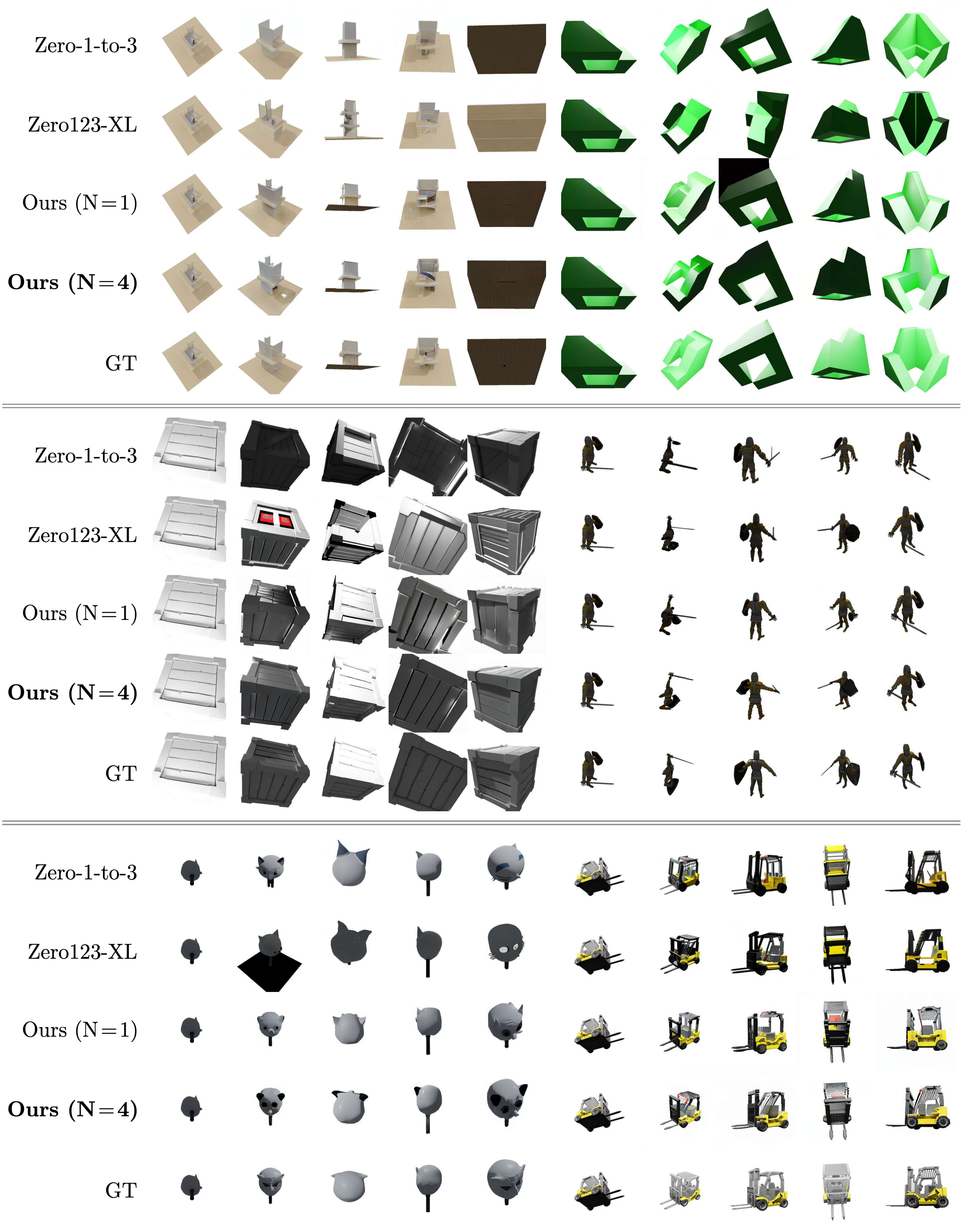}
    \caption{Novel view synthesis examples by Zero-1-to-3, Zero123-XL, and our NVS-Adapter with $N=1$ and $N=4$ on Objaverse.}
     \label{fig:supp_qual_objaverse}
    \vspace{-0.1in}
\end{figure}

\begin{figure}
    \centering
    \includegraphics[width=\textwidth]{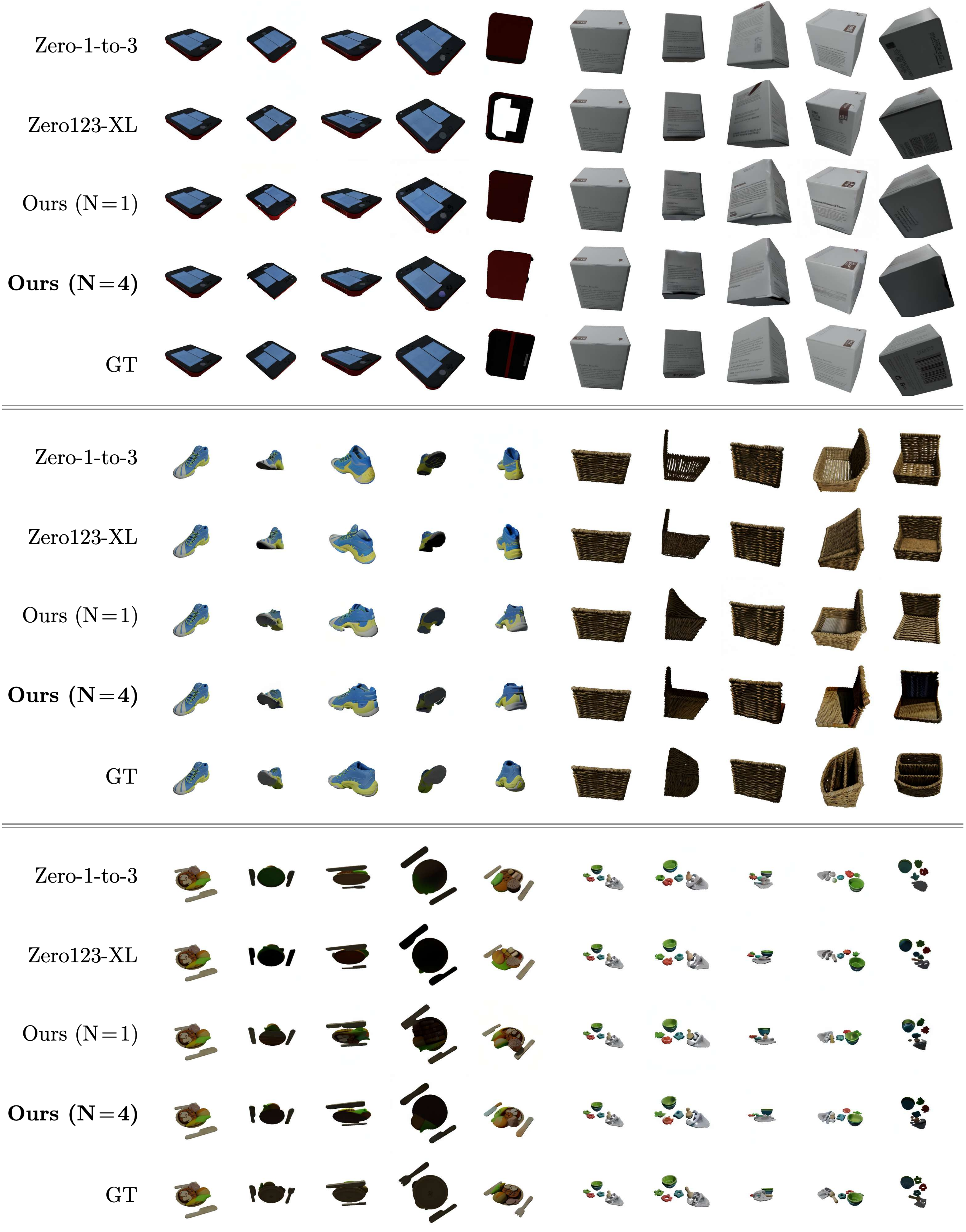}
    \caption{Novel view synthesis examples by Zero-1-to-3, Zero123-XL, and our NVS-Adapter with $N=1$ and $N=4$ on GSO.}
     \label{fig:supp_qual_gso}
    \vspace{-0.1in}
\end{figure}
\begin{figure}
    \centering
    \includegraphics[width=\textwidth]{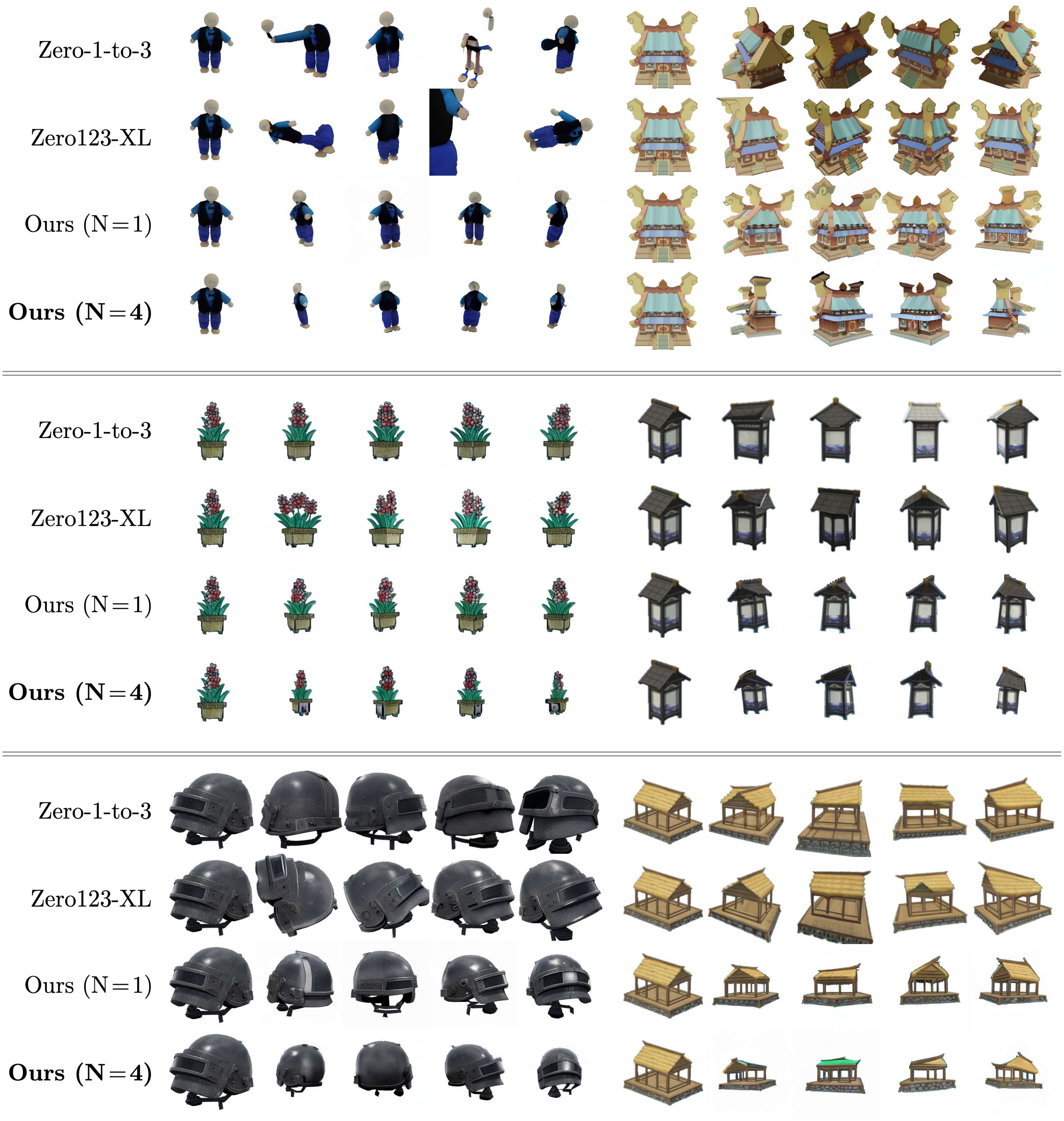}
    \caption{Novel view synthesis examples by Zero-1-to-3, Zero123-XL, and our NVS-Adapter with $N=1$ and $N=4$ on reference images from the internet.}
     \label{fig:supp_qual_inthewild}
    \vspace{-0.1in}
\end{figure}

\begin{figure}
    \centering
    \includegraphics[width=0.8\textwidth]{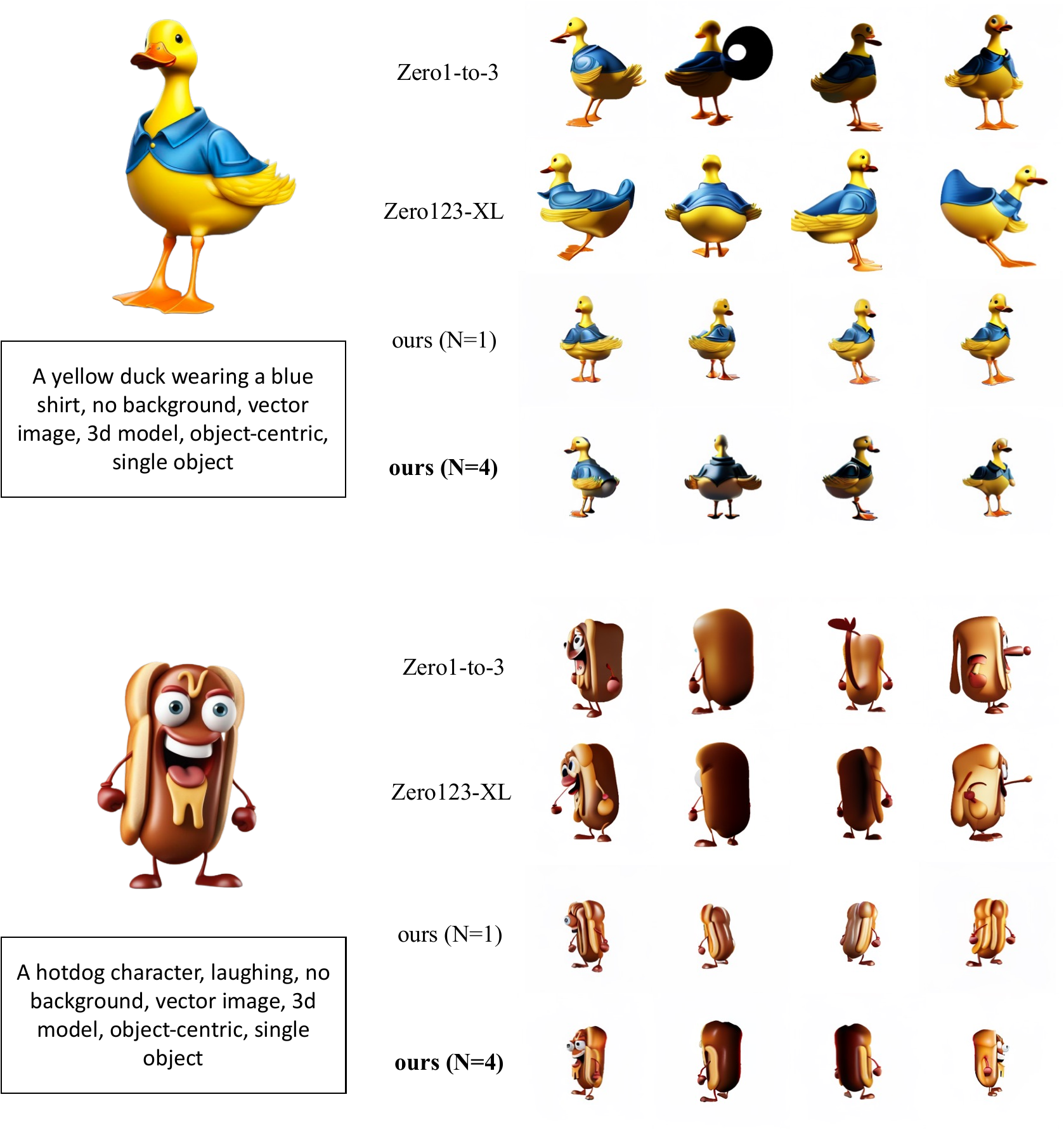}
    \caption{Novel view synthesis examples by Zero-1-to-3, Zero123-XL, and our NVS-Adapter with $N=1$ and $N=4$ on images generated from text prompts.}
     \label{fig:supp_qual_textprompt1}
    \vspace{-0.1in}
\end{figure}

\begin{figure}
    \centering
    \includegraphics[width=0.8\textwidth]{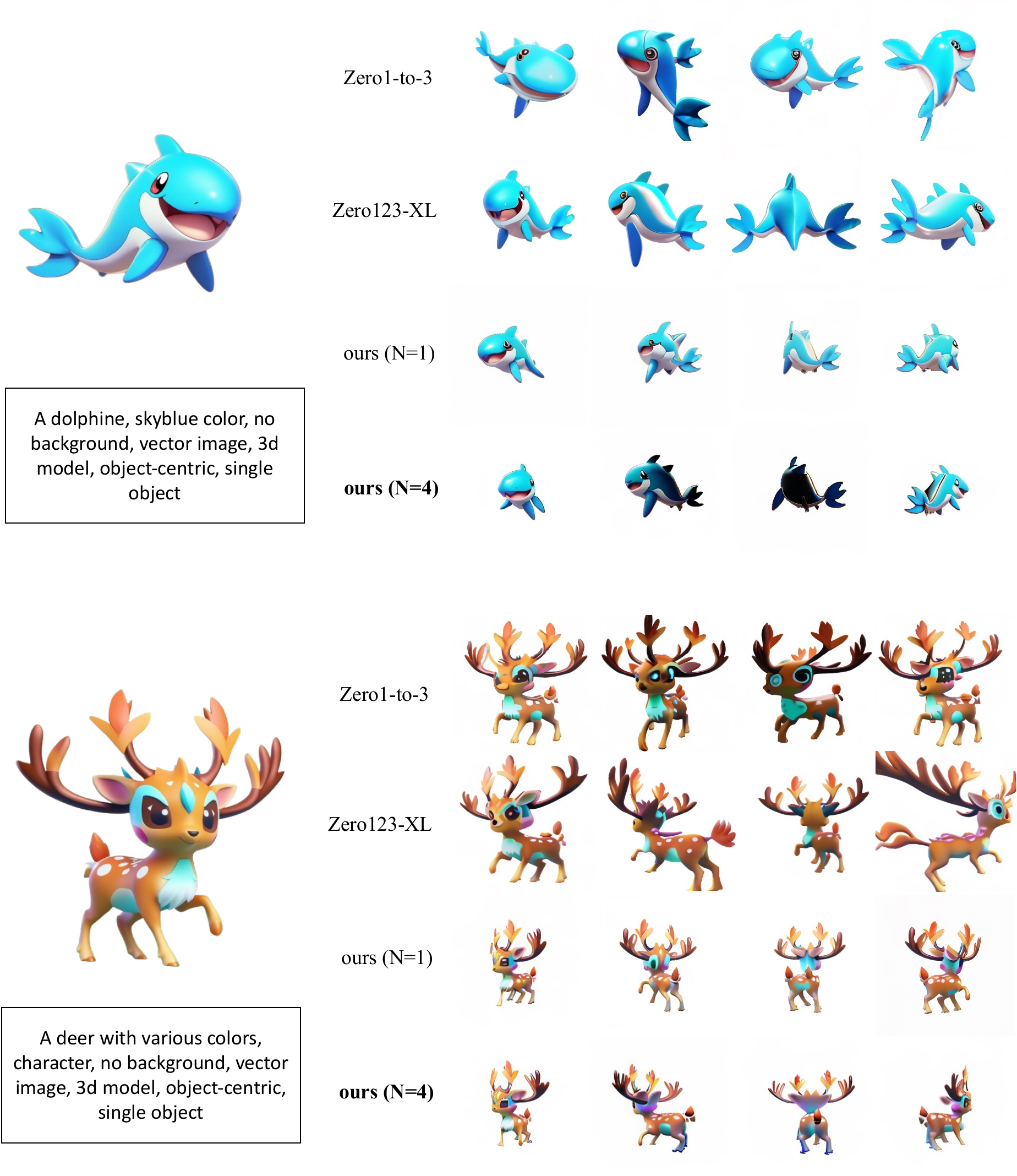}
    \caption{Novel view synthesis examples by Zero-1-to-3, Zero123-XL, and our NVS-Adapter with $N=1$ and $N=4$ on images generated from text prompts.}
     \label{fig:supp_qual_textprompt2}
    \vspace{-0.1in}
\end{figure}

\begin{figure}
    \centering
    \includegraphics[width=\textwidth]{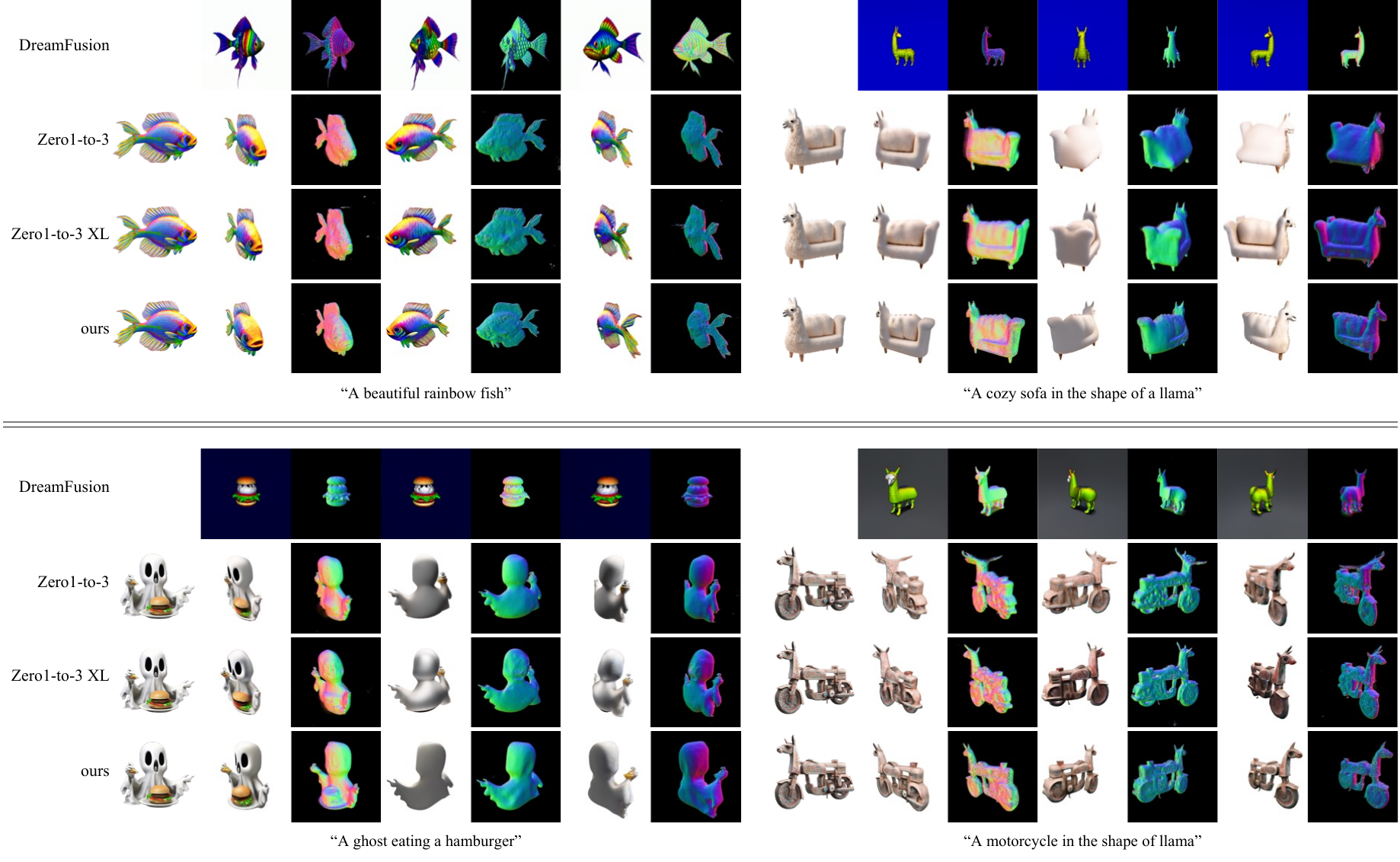}
    \caption{3D reconstruction examples via Score Distillation Sampling (SDS) with baselines and our NVS-Adapter. Images shows 3D reconstructions results conditioned on an image generated by T2I model.}
     \label{fig:supp_qual_sds_text}
    \vspace{-0.1in}
\end{figure}

\begin{figure}
    \centering
    \includegraphics[width=\textwidth]{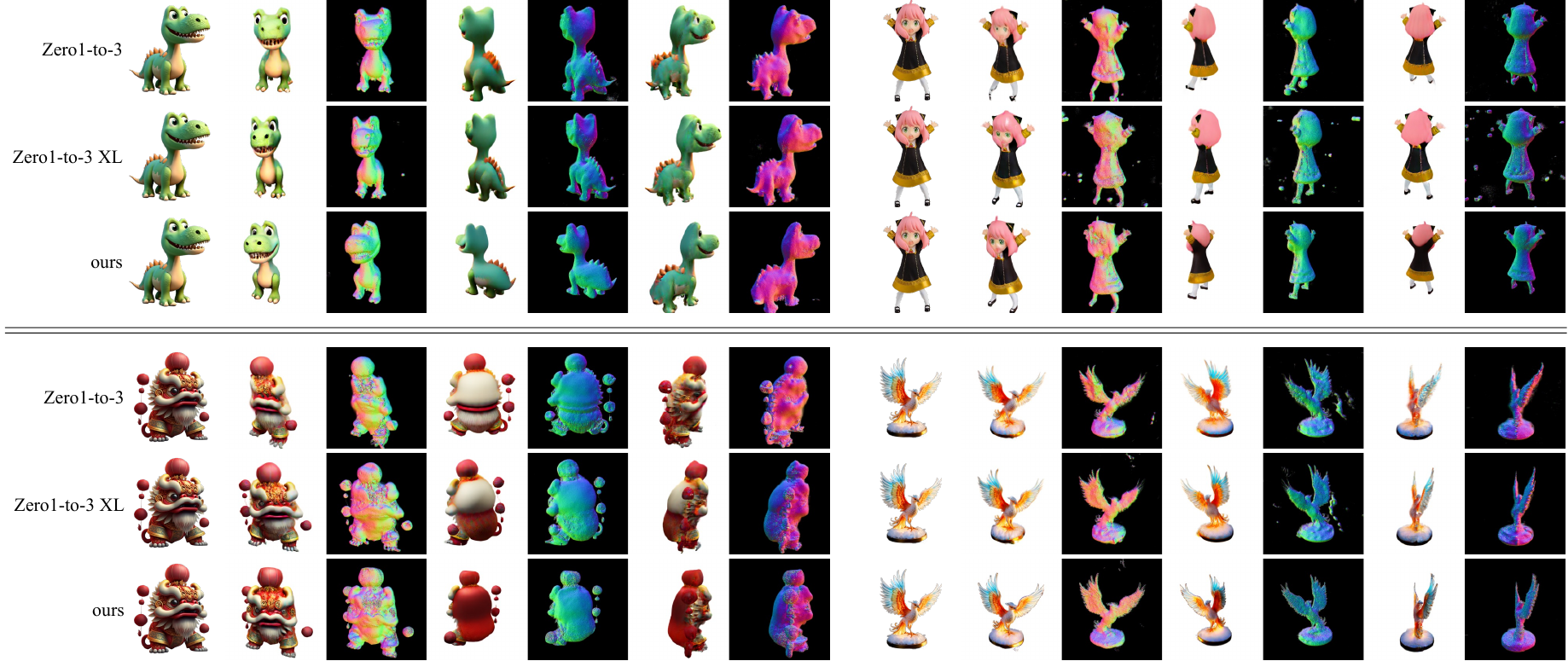}
    \caption{3D reconstruction examples via Score Distillation Sampling (SDS) with baselines and our NVS-Adapter. Images shows 3D reconstructions results on an image from the internet.}
     \label{fig:supp_qual_sds_image}
    \vspace{-0.1in}
\end{figure}

\begin{figure}
    \centering
    \includegraphics[width=0.8\textwidth,]{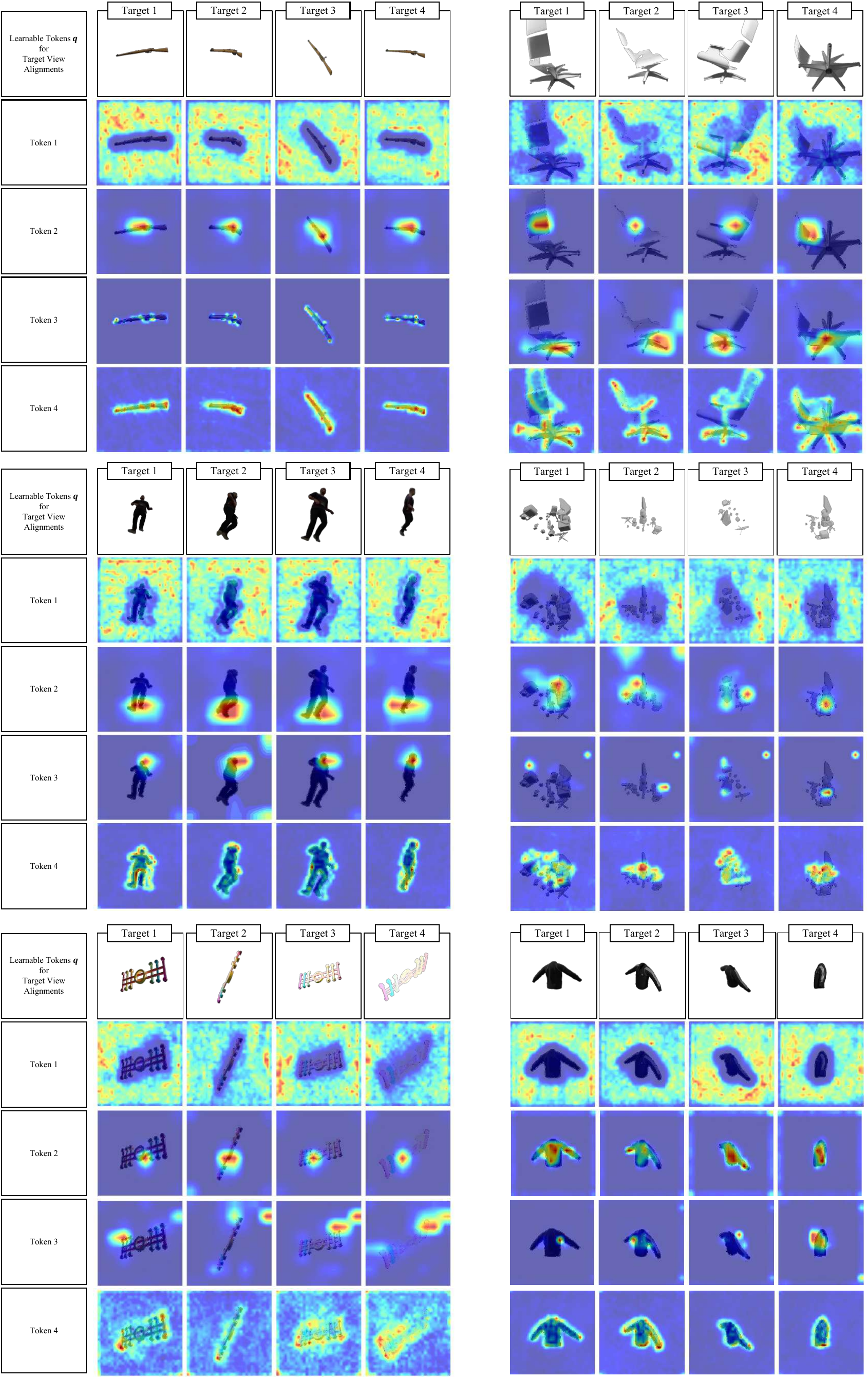}
    \caption{Visualization of attention maps of target views alignment. Each token aggregates different regions of target views.}
     \label{fig:supp_qual_target_attn}
    \vspace{-0.1in}
\end{figure}

\begin{figure}
    \centering
    \includegraphics[width=0.8\textwidth]{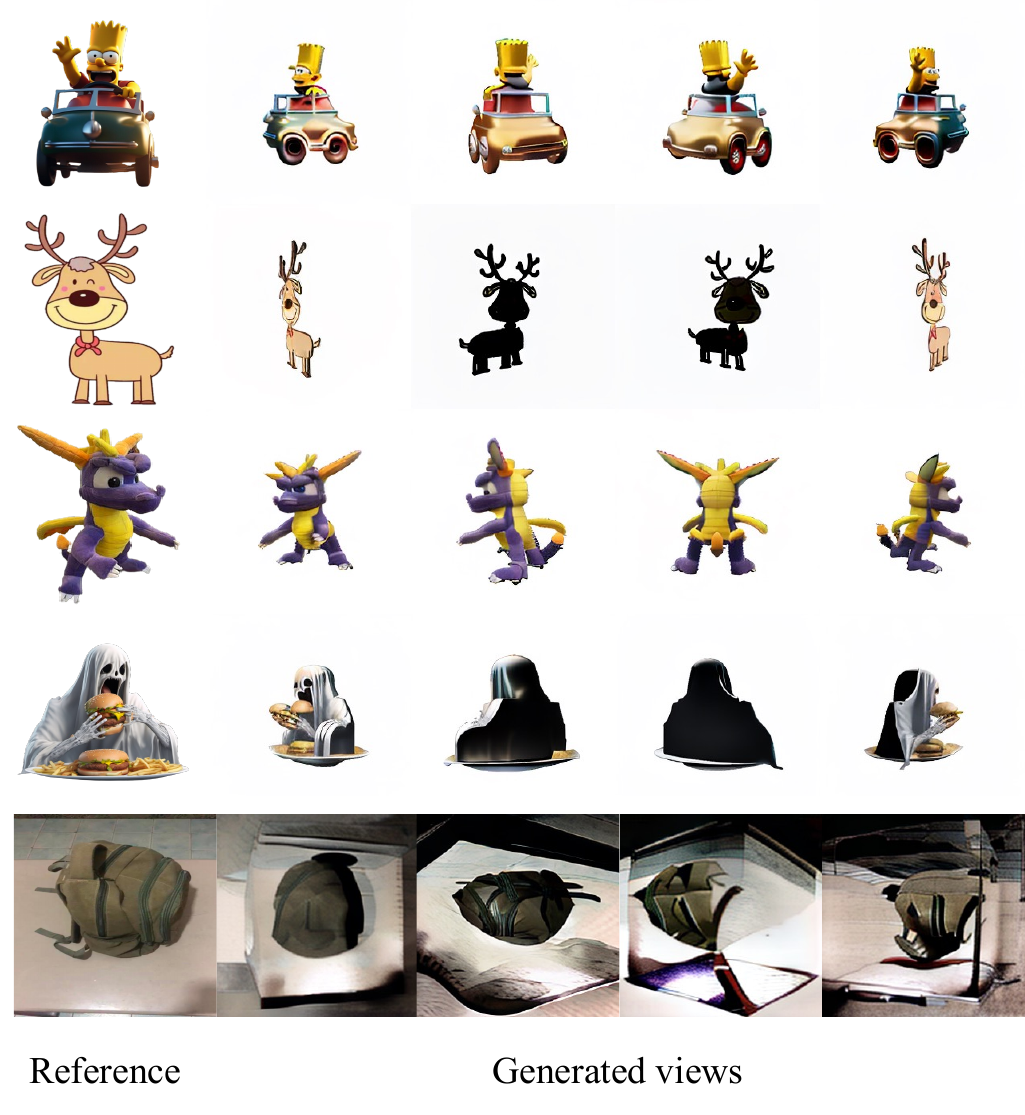}
    \caption{Several failure cases of our NVS-Adapter. The first column shows reference images and the rest of the columns are rendered novel views. }
     \label{fig:supp_qual_failure}
    \vspace{-0.1in}
\end{figure}